\documentclass[11pt,a4paper]{article}
\usepackage[hyperref]{acl2020}
\usepackage{times}
\usepackage{latexsym}

\usepackage{amsfonts}
\usepackage{amsmath}
\usepackage{booktabs}
\usepackage{graphicx}
\usepackage{caption}
\usepackage{subcaption}
\usepackage{enumitem}

\usepackage{microtype}

\aclfinalcopy % Uncomment this line for the final submission
\def\aclpaperid{1017} %  Enter the acl Paper ID here

\title{Highway Transformer: Self-Gating Enhanced Self-Attentive Networks}
\author{Yekun Chai$^\dagger$ \quad Shuo Jin$^\ddagger$ \quad Xinwen Hou$^\dagger$ \\
  $^\dagger$Institute of Automation, Chinese Academy of Sciences \\
  $^\ddagger$University of Pittsburgh \\
    \texttt{chaiyekun@gmail.com} 
    \, \texttt{shj42@pitt.edu}
  }

\date{}

\begin{document}
\maketitle
\begin{abstract}
Self-attention mechanisms have made striking state-of-the-art (SOTA) progress in various sequence learning tasks, standing on the multi-headed dot product attention by attending to all the global contexts at different locations. Through a \emph{pseudo information highway}, we introduce a gated component \emph{self-dependency units} (SDU) that incorporates LSTM-styled gating units to replenish internal semantic importance within the multi-dimensional latent space of individual representations. The subsidiary content-based SDU gates allow for the information flow of modulated latent embeddings through skipped connections, leading to a clear margin of convergence speed with gradient descent algorithms. We may unveil the role of gating mechanism to aid in the context-based Transformer modules, with hypothesizing that SDU gates, especially on shallow layers, could push it faster to step towards suboptimal points during the optimization process.
% the result of xl?
% scalability
\end{abstract}

\section{Introduction}
Self-attention mechanism has lately attracted extensive interests due to its remarkable achievement on a wide range of sequence modeling applications, including natural language processing such as neural machine translation~\citep{vaswani2017attention, ott2018scaling, shaw2018self}, language modeling (LM)~\citep{dai2019transformer,al2019character}, self-supervised pretraining~\citep{radford2018improving,devlin2018bert,lan2019albert}; image generation~\citep{parmar2018image}; deep reinforcement learning~\citep{zambaldi2018deep, vinyals2019grandmaster}, \emph{etc}. 

Holding the great promise of deep neural networks in language and images, Transformer capitalizes on the stacked multi-headed self-attention mechanism based on the conventional encoder-decoder architecture in a sequence-to-sequence (seq2seq) manner to learn the global soft signals without explicit recurrence mechanism. Multi-head dot product attention (MHDPA) not only underpins the parallel training of multiple heads but captures long-term dependencies across an arbitrarily long distance within the same context. In which separated multiple heads independently draw sub-level attentions within the latent semantic sub-space of a fixed dimension, where different heads are presumed to signal different meaning aspects implicitly~\citep{vaswani2017attention}. Additionally, residual connections between layers allow the deep tandem stack of multiple identical modules by impeding \emph{degradation} problem during training~\citep{he2016deep}. Thus Transformer architectures take the place of Recurrent Neural Networks (RNNs), especially Long Short-Term Memory (LSTM) networks~\citep{hochreiter1997long} to be the model solution to learning sequential data.

% MADPA 
% in parallel-> faster 
% residual conn / information flow

\begin{figure}[]
% \vskip 0mm
\begin{center}
\includegraphics[width=\columnwidth]{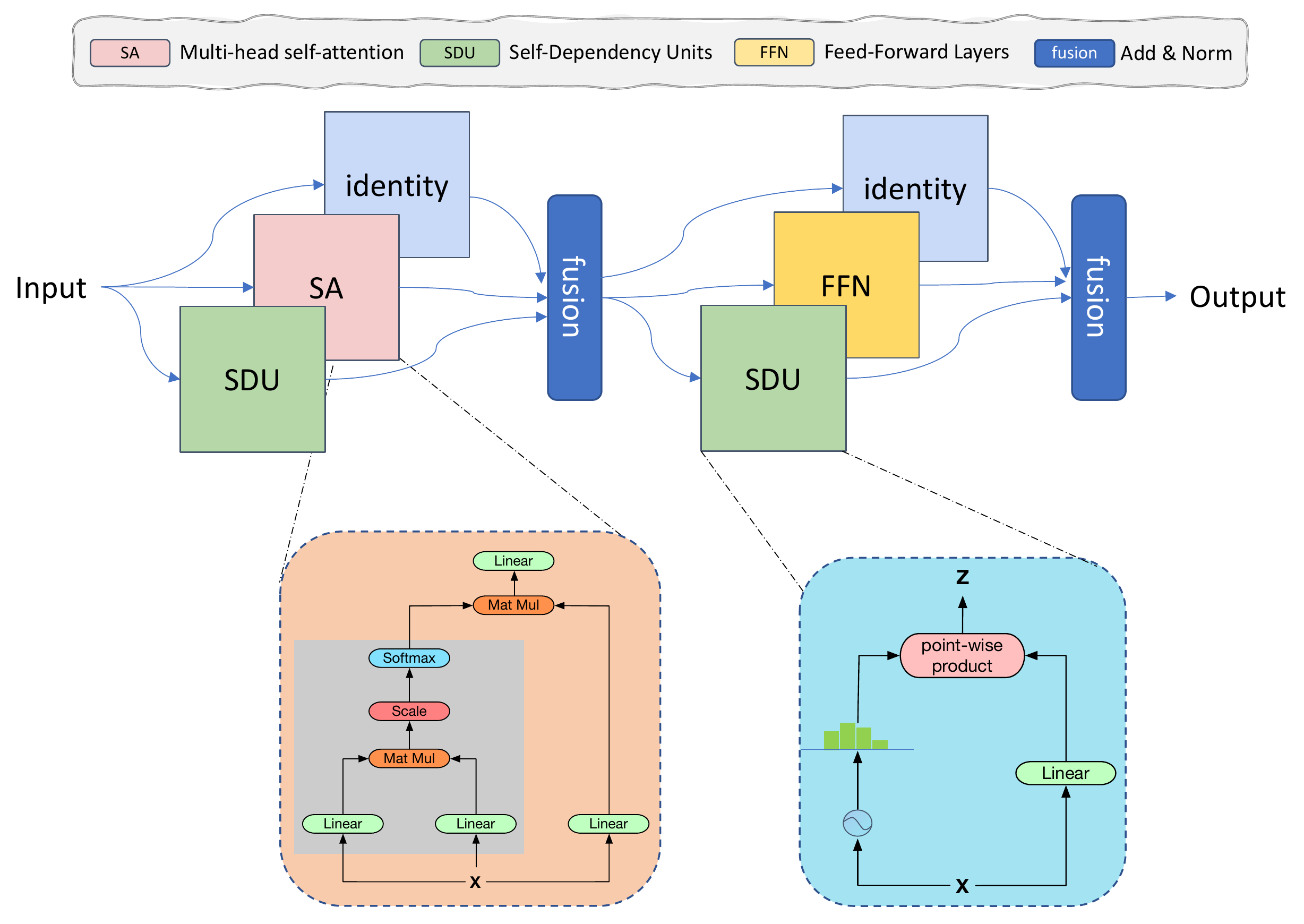}
% \vskip -3mm
\caption{Illustration of highway Transformer.}
\label{fig:highway_trm}
\end{center}
\vskip -7mm
\end{figure} 

Recently, there have been plenty of works contending that gating mechanisms could play a vital role or even entirely substitute RNNs or Transformers to model language sequences. \citet{dauphin2017language} firstly claimed that non-recurrent networks are also highly competitive with conventional RNN-dominated models in LM. They proposed the hierarchical gated temporal convolution neural networks (CNNs) with Gated Linear Units (GLU) to replace the recurrent connections in RNNs and achieved strong performance with faster training speed. \citet{gehring2017convolutional} integrated absolute positional embedding, multi-step attention, GLU, and residual connections into entirely convolutional models to outperform strong LSTM models in NMT and abstractive summarization tasks. \citet{wu2019pay} applied dynamic convolutions using shared softmax-normalized filters of depth-wise on GLU-regulated inputs within a fixed reception field rather than global contexts, challenging the common self-attention-dominated intuition. 

However, all of the models, as mentioned earlier, adopt stacked CNNs rather than self-attention networks (SAN) to attend to the global contexts. It is well-known that CNNs are good at learning local-region features rather than long-term dependency, while SANs are adept in attending global dependencies. Context-based self-attention can capture the importance of relative relations under a valid context and is thus location-unaware. It focuses on the object-wise attention distributions between any two words but ignores the fundamental importance of feature-wise information. 

Intuitionally, people need to consider not only the global contextual dependency but the meaning of individual words to comprehend the reading materials better. Grounding on this, we apply self-gating approaches on Transformer blocks for seq2seq modeling that combines gating units with skip-connections and Transformers to jointly take into account both the inner feature-wise importance and the relation-aware content-based attention distribution. 

We adopt the self-dependency gating approach to intrinsically draw a binary importance ratio of itself and decide how much information of each feature to retain or remove. 
% We probe whether such gated units can benefit for the Transformer and its variant architectures and empirically prove that it is usually beneficial for the convergence speed during training and could achieve obvious performance on shallow Transformer components, substantiating the argument that bottom layers of SAN can learn more on local contexts~\citep{yang2018modeling}.
Our key contributions are:
\setlist{nolistsep}
\begin{itemize} [noitemsep]
    \item to illustrate that our self-dependency units on shallow Transformer layers could expedite the convergence speed during both the training and validation process without hyperparameter tuning.
    \item to support the claim that Transformer layers in different depth attend to information of different aspects, wherein bottom layers focus on local-range encodings. It substantiates the argument that the bottom layers of SAN can learn more in local contexts~\citep{yang2018modeling}. 
    \item to empirically prove that self-gating mechanisms are complementary to recurrence mechanisms in R-Transformer and Transformer-XL components.

\end{itemize}{}

% support the common sense that gated models usually do not have too many stacked layers

%  Inspired by 1) highway information flow, 2) gated CNN / pay less / context CNN.

% content-based gate / bear a resemblance to channel-wise location-based attention
% argue that long dependencies rather than the inner information
% context-based attention / location-irrelevant
% Relational (Relative attention) Attn -> relation between each two objects. -> relatively attentive, focus on the relation pairs, rather than each embedding dim inside. We draw an internal/intrinsic binary attention distribution that it is also important to capture the inner distribution among each word. People do not always pay attention after looking through all the words, inversely they usually could determine the importance of each sequence at the first glare by merely looking at itself, like focusing on some entities and relations when do reading comprehension tasks other than trivial stopwords.

% illustrate the gated CNN can serve in LMs
% model internal semantic correlation!

% contribution
% 1. combine the lstm-styled gate by extending Highway connection
% 2. better convergence / good scalability with few line code
% 3. also indicate the intuition of the attention distribution
% 4. draw attention when they first see it. Interior attention only based on its features.
% 5. fine-granularity of the  using Bernoulli distribution

% Self-gating can be seen as an importance ratio with fine-granularity.

\section{Preliminaries}
\label{sec:pre}

This section briefly introduces the related background of Transformer and Highway Networks.

% MHDPA
SAN has been dominant in most SOTA sequence learning models, whose basic components consist of stacked Transformers modules. We conduct comparison experiments on the Transformer and its two variants, Transformer-XL~\citep{dai2019transformer} and R-Transformer~\citep{wang2019r}.
\subsection{Multi-head Dot Product Attention}
% \paragraph{Dot product self-attention} 
% DPA
% mhdpa
Scaled dot product attention (DPA)~\citep{vaswani2017attention} computes global attention weights between pairs within the context across an arbitrarily long distance, which could allow the simultaneous training and space-saving, impeding the drawbacks of sequential dependency of RNNs.

Given the input word representation $\mathbf{X} \in \mathbb{R}^{L \times dh}$, where $L$ is the sequence length, $d$ is the input dimension of each head and $h$ is the number of attention heads, DPA uses the linear projection to acquire the query $\mathbf{Q}$, key $\mathbf{K}$ and value $\mathbf{V}$. Denoting splitted inputs for $i$-th head as $\mathbf{X}_i \in \mathbb{R}^{L \times d}$, where $i \in \{1, \cdots, h\}$, single-head self-attention can be calculated as:
\begin{align}
\begin{split}\label{eq:qkv}
    \mathbf{Q}_i,\mathbf{K}_i,\mathbf{V}_i ={}& \mathbf{X}_i\mathbf{W}_q, \mathbf{X}_i\mathbf{W}_k, \mathbf{X}_i\mathbf{W}_v
\end{split}\\
\begin{split}\label{eq:dpa}
    \textrm{head}_i ={}& \text{softmax}\left( d^{-1/2}\mathbf{Q}_i \mathbf{K}_i^\top \right) \mathbf{V}_i
\end{split}
\end{align}{where learnable weights $\{\mathbf{W}_q, \mathbf{W}_k,\mathbf{W}_v \} \in \mathbb{R}^{d \times d}$, $d^{-1/2}$ is a scaling factor to prevent the effect of large values. In LM tasks, attention weights before softmax function are masked to only attend to history sequences.}

% \begin{equation}
% \label{eq:dpa}
%     \textit{DPA}(\mathbf{x}) = \text{softmax}\left( \frac{\mathbf{x}\mathbf{W_q} \mathbf{W_k^\top} \mathbf{x^\top}}{\sqrt{d}} \right) \mathbf{x}\mathbf{W_v}
% \end{equation}

MHDPA (Fig~\ref{fig:san}) linearly projects the single DPA into $h$ heads and performs attention operation in parallel, to jointly learn different semantic meanings of different subspaces~\citep{vaswani2017attention}. MHDPA can be calculated as:
\begin{equation}
\label{eq:mhdpa}
    \textrm{Att}(\mathbf{Q},\mathbf{K},\mathbf{V}) = \left[\text{head}_1 \circ \cdots \circ \text{head}_h \right] \mathbf{W_o}
\end{equation}{where $\circ$ denotes the concatenation of $h$ different heads, $\mathbf{W_o} \in \mathbb{R}^{dh \times dh}$ is the trainable weight}.

\subsection{Transformer}
\paragraph{Absolute Positional Encoding}
Transformer applies sinusoidal timing signal as the absolute positional encoding (PE) and directly element-wise add the dense word embeddings $\mathbf{E} \in \mathbb{R}^{L \times dh}$ on the PE before feeding into Transformer modules:
\begin{align}
\begin{split}\label{eq:sin-pe}
    PE_{(\textrm{pos}, 2i)} ={}& \sin(\frac{\textrm{pos}}{10000^{2i/d}}) 
\end{split}\\
\begin{split}\label{eq:cos-pe}
    PE_{(\textrm{pos}, 2i+1)} ={}& \cos(\frac{\textrm{pos}}{10000^{2i/d}})
\end{split}\\
\begin{split}\label{eq:pe}
    \mathbf{X} ={}& \mathbf{E} + PE(\mathbf{E})
\end{split}
\end{align}{where `$\textrm{pos}$' indicates the position of sequences, $i$ denotes the order along the embedding dimension.}

Given input representations $\mathbf{X}$, Transformer components with a sternward Layer Normalization (LN) is:
\begin{align}
\begin{split}
    \mathbf{U} ={}& LN(\mathbf{X} + \textrm{Att}(\mathbf{Q},\mathbf{K},\mathbf{V}) \label{eq:att_m}
\end{split}\\
\begin{split}\label{eq:FFN}
    \textrm{FFN}(\mathbf{U}) ={}& \textit{FF} \big( \text{ReLU}(\textit{FF}(\mathbf{U})) \big)
\end{split}\\
    \mathbf{O} ={}& LN(\mathbf{U} + \textrm{FFN}(\mathbf{U})) \label{eq:fc_m}
\end{align}{where Eq.~\ref{eq:FFN} indicates the position-wise feed-forward networks (FFN), $\mathbf{O} \in \mathbb{R}^{L \times dh}$ represents the output of transformer layer. \textit{FF} denotes the feed-forward fully-connected layer, ReLU is used as the non-linear activate function.}

\subsection{Transformer-XL}
Transformer-XL~\citep{dai2019transformer} injected relative PE and segment-level recurrence to provide historical information for LM tasks.
\paragraph{Relative Positional Encoding}
Transformer-XL decomposed the dot product calculation of MHDPA, merged terms with similar meanings of positional bias, and reduced trainable weights with global positional semantics. It incorporated partial trainable parameters of relative sinusoidal PE in the MHDPA operation.

The Relative PE $A^\text{rel}$ of Transformer-XL is:
\begin{align}
\begin{split}\label{eq:a}
    a ={}& \mathbf{Q}^\top \mathbf{K}
\end{split}\\
\begin{split}\label{eq:b}
    b ={}& \mathbf{Q}^\top \mathbf{W_{k,R}} \, \mathbf{R}
\end{split}\\
\begin{split}\label{eq:c}
    c ={}& \mathbf{u}^\top \, \mathbf{K}
\end{split}\\
\begin{split}\label{eq:d}
    d ={}& \mathbf{v}^\top \mathbf{W_{k,R}}\, \mathbf{R}
\end{split}\\
    A^\text{rel} (\mathbf{Q}, \mathbf{K}) ={}& a+b+c+d \label{eq:e}
\end{align}{where $ \mathbf{W_{k,R}} \in \mathbb{R}^{d \times d}$, $\{\mathbf{u},\mathbf{v}\} \in \mathbb{R}^{d}$  are trainable parameters. For each two positions $i,j$ in the segment, $\mathbf{R}$ is sinusoidal encoding matrices between relative position $i-j$. The terms $a, b, c, d$ in the Eq.~\ref{eq:a},~\ref{eq:b},~\ref{eq:c},~\ref{eq:d} represent the content-based addressing, content-dependent positional biases, global biases between different positions and the global positional biases, respectively.
}

% \begin{align}
% \begin{split}\label{eq:xl-rpe}
%     A_{i,j}^\text{rel} ={}& \mathbf{x}_{i,\bullet}^\top W_q^\top W_{k,E} \mathbf{x}_{\bullet,j} \\ &+ \mathbf{x}_{i,\bullet}^\top W_q^\top W_{k,R}  R_{i-j} \\
%   & + u^\top W_{k,E} \mathbf{x}_{\bullet,j} \\ &+ v^\top W_{k,R} R_{i-j}
% \end{split}
% \end{align}

\paragraph{Segment-level Recurrence}
In Transformer-XL, the previous hidden states are cached and reused to inject the history information and attend to contexts beyond a fixed length through multi-layer stacks. 
The MHDPA is computed as:
\begin{align}
\begin{split}
    \mathbf{M}_{\tau}^{n-1} ={}& \overbrace{SG(\mathbf{X_{\tau-1}^{n-1}})}^\text{stop gradient} \circ \, \mathbf{X_{\tau}^{n-1}}
    \label{eq:xl-k-sg}
\end{split}\\
\begin{split}
   \mathbf{Q}, \textbf{K}, \mathbf{V} ={}& \mathbf{X_{\tau}^{n-1}}\mathbf{W_q}, \,\mathbf{M}_{\tau}^{n-1}\mathbf{W_k}, \, \mathbf{M}_{\tau}^{n-1}\mathbf{W_v} \label{eq:xl-qv}
\end{split}\\
    DPA &(\mathbf{Q}, \textbf{K}, \mathbf{V} ) = A^\text{rel} (\mathbf{Q}, \mathbf{K}) \mathbf{V} \label{eq:xl-eq}
\end{align}{wherein the key and value $\mathbf{M}_{\tau}^{n-1}$ concatenate the previous memory $\mathbf{X_{\tau-1}^{n-1}}$ with the current segment inputs $\mathbf{X_{\tau}^{n-1}}$ for the $\tau$-th segment in the $n$-th layer, $SG$ means no backpropagation through the tensor.}

\subsection{R-Transformer}
% lcoal RNN
R-Transformer~\citep{wang2019r} employed short-range RNNs, termed \textit{localRNNs}, to capture the positional information without explicit PEs. \textit{localRNNs} take the recurrent connections within a local context, and shift right with one position at each time step. It can be seen as applying the RNN cells, such as LSTM, on the same receptive fields as the convolutional filters along the sequence direction.
\begin{align}
\begin{split}\label{eq:r-t0}
    \mathbf{X} ={}& \textrm{localRNN}(\mathbf{E})
\end{split}\\
\begin{split}\label{eq:r-t1}
    \mathbf{O} ={}& \textrm{Transformer-layer}(\mathbf{X})
\end{split}
\end{align}

None of the above Transformer models explicitly consider the essential feature-wise information. We augment several gated units on the Transformer block of the models above and empirically illustrate the effectiveness of gating units on convergence acceleration.

\begin{figure*}[t]
% \vskip -5mm
     \centering
     \begin{subfigure}[b]{0.35\textwidth}
         \centering
         \includegraphics[width=\textwidth]{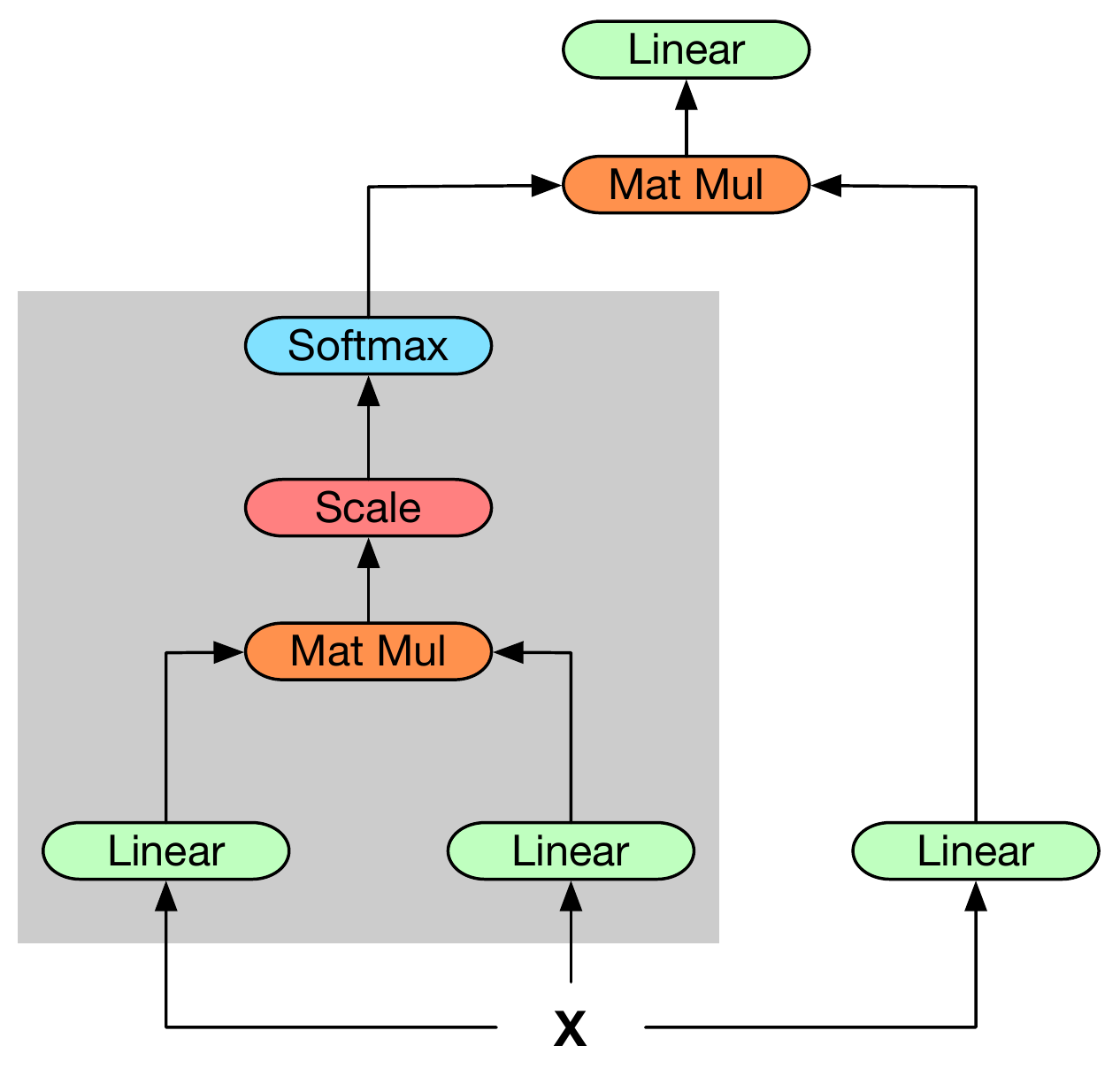}
         \caption{Multi-head dot product self-attention}
         \label{fig:san}
     \end{subfigure}
     \hfill
     \begin{subfigure}[b]{0.25\textwidth}
         \centering
         \includegraphics[width=\textwidth]{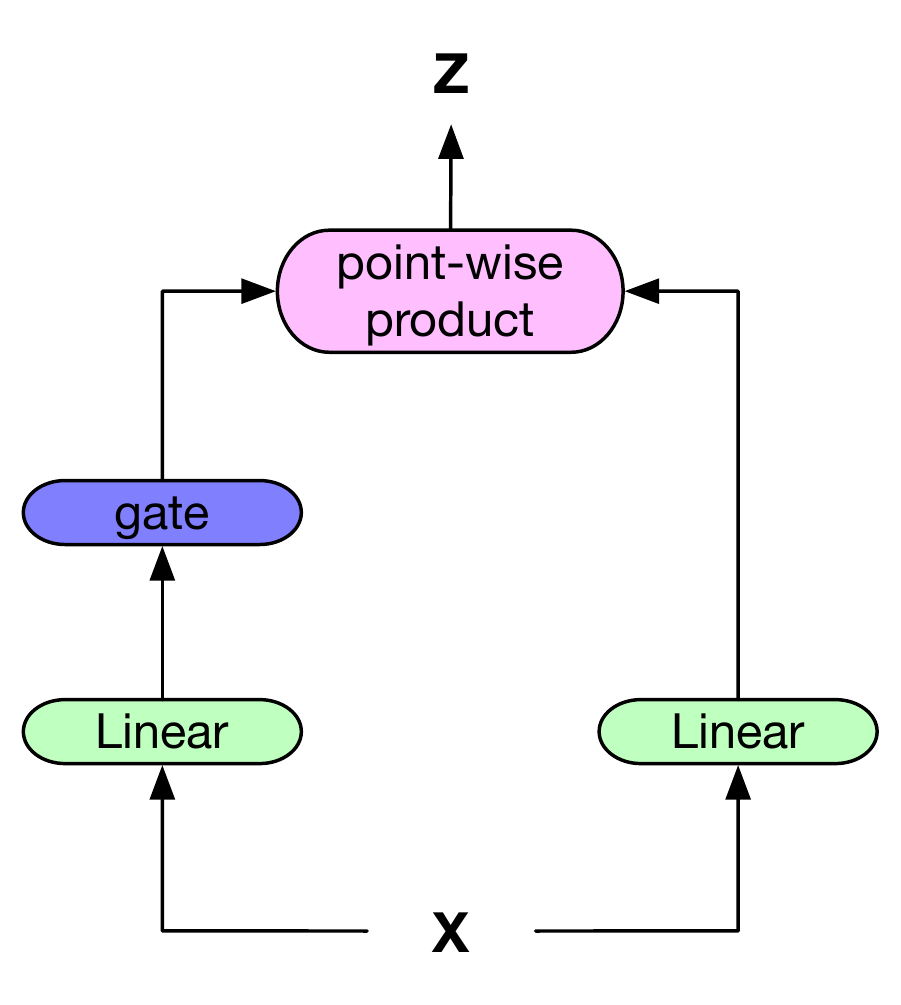}
         \caption{Self-dependency units}
         \label{fig:sdu}
     \end{subfigure}
     \hfill
     \begin{subfigure}[b]{0.35\textwidth}
         \centering
         \includegraphics[width=\textwidth]{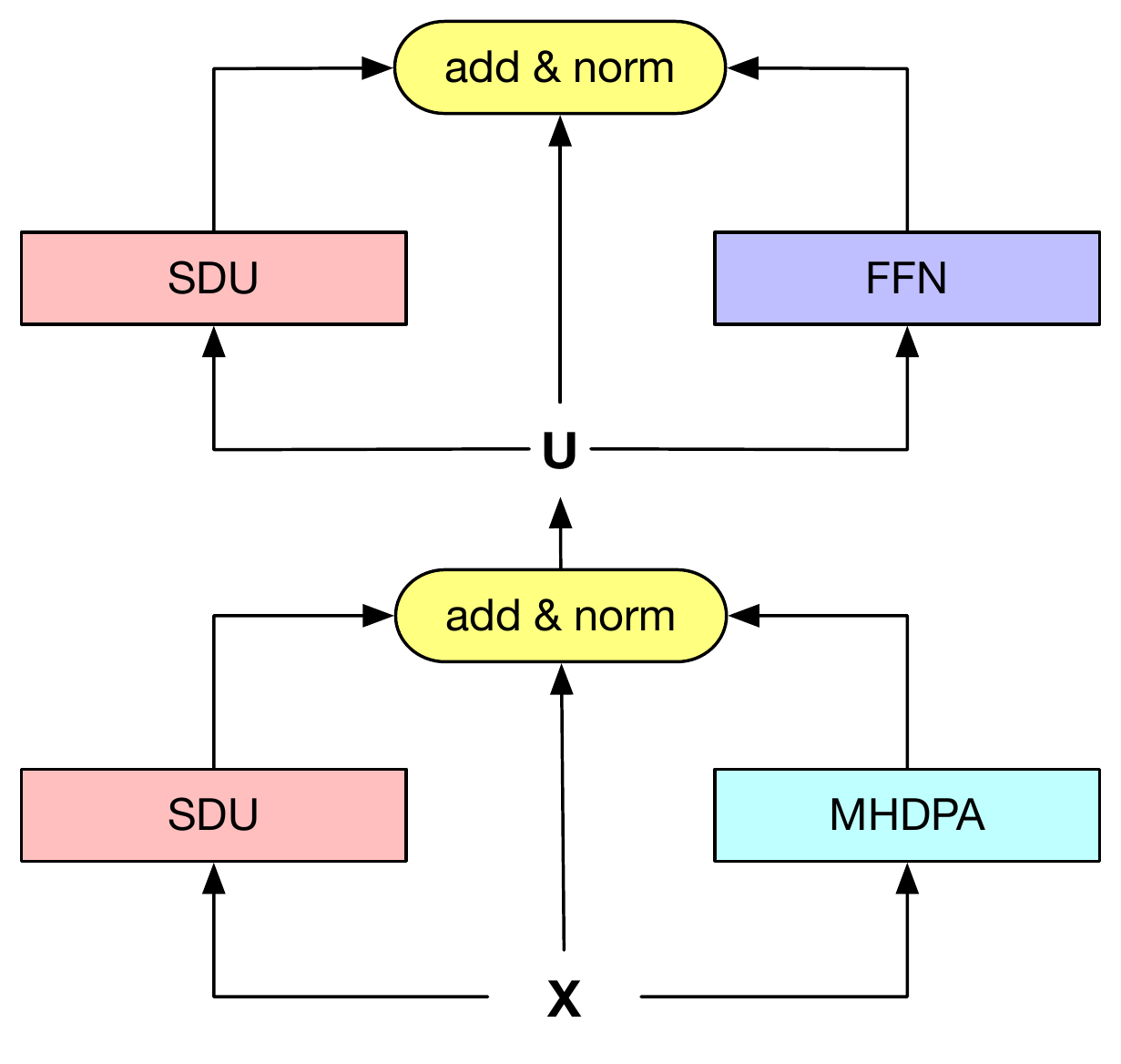}
         \caption{SDU-augmented Transformer block}
         \label{fig:block}
     \end{subfigure} \vskip -2mm
        \caption{Illustration of MHDPA, SDU and SDU-enhanced Transformer block.}
        \label{fig:three graphs}
        \vskip -4mm
\end{figure*}

\subsection{Highway Networks}
Let we define the non-linear transforms as $H$, $T$ and $C$, Highway Network~\cite{srivastava2015highway} is defined as:
\begin{equation}
    \mathbf{O} = H(\mathbf{X})\odot T(\mathbf{X}) + \mathbf{X}\odot  C(\mathbf{X}) 
\end{equation}{where $T(\cdot)$ and $C(\cdot)$ denote transform and carry gates to control the input transformation, $\odot$ denotes the Hadamard product. 
}

% model
\section{Gating Architecture}
LSTM-styled gate units have been proven to be effective on sequence learning tasks~\citep{dauphin2017language,gehring2017convolutional,wu2019pay}. We spontaneously wonder whether such gating mechanisms could help when augmenting the Transformer components. 
% LSTM
\subsection{Self-Dependency Units}
Similar to GLU~\citep{dauphin2017language} that adopts the inputs as sigmoidal gates, we apply the Self-Dependency Units (SDU) by taking full inputs as their respective self gates and computing the element-wise product upon themselves (Fig~\ref{fig:sdu}).
\begin{align}
\begin{split} \label{eq:sdu-gate}
    T(\mathbf{X}) ={}& \mathbf{\Psi}(\mathbf{X}\mathbf{W_1}+\mathbf{b_1})
\end{split}\\
    \textrm{SDU}(\mathbf{X}) ={}& T(\mathbf{X}) \odot (\mathbf{X}\mathbf{W_2}+\mathbf{b_2}) \label{eq:sdu}
\end{align}{where $T(\mathbf{X})$ indicates the \emph{transform} gate, $\mathbf{\Psi}$ is the gate function that confine the linear projection into a fixed range, $\{\mathbf{W_1}, \mathbf{W_2}\} \in \mathbb{R}^{d \times d}$ and $\{\mathbf{b_1}, \mathbf{b_2}\} \in \mathbb{R}^d$ are trainable parameters.}

 The element-wise gating function $\mathbf{\Psi}$ takes sigmoidal-curve functions to regulate the point-wise weights within a fixed region, which have a side effect of relative normalization. Specifically, the sigmoid function $\sigma(x)=1/(1+\exp(-x))$ and its rescaled version $\tanh(x) = 2\sigma(2x)-1, \textrm{where } x \in \mathbb{R}$. 
 
 We interpret the \emph{tanh} function as an update gate, which can restrict the importance range into between -1 and 1, while the $\sigma$ function bears a resemblance to the input gate in LSTMs to modulate how much information to retain at the feature-wise level.

 \subsection{Pseudo-highway Connection}
MHDPA computes the multi-headed pairwise attention along the sequence dimension by measuring the distance between each word. It might overlook the fundamental importance of individual features. Rather than replacing MHDPA as gating and convolution operations in dynamic convolutions~\citep{wu2019pay}, we simply add a new branch of inputs to enrich the representations of residual connected MHDPA with augmented gating-modified encodings. The gated units are also supplemented on FFN modules to provide additional self-adaptive information flow ( Fig~\ref{fig:block}).

From other perspectives, SDU can be considered as a self-dependency non-linear activation function with dynamic adaptation. The self-gating augmented Transformer module is calculated as:
\begin{align}
\begin{split}
    \mathbf{U} ={}& LN \big(\mathbf{X} + \textrm{Att}(\mathbf{Q},\mathbf{K},\mathbf{V}) \\& + \textrm{SDU}(\mathbf{X}) \big) \textbf{\label{eq:t1}}
\end{split}\\
    \mathbf{O} ={}& LN \big(\mathbf{U} + \textrm{FFN}(\mathbf{U})  + \textrm{SDU}(\mathbf{U}) \big) \label{eq:t2}
\end{align}{where $\mathbf{U}$ and $\mathbf{O}$ represent the intermediate representation and outputs.}

\paragraph{Pseudo-highway Transformer} When we take $\sigma$ gate as $\Psi$, we can have the similar format as highway networks:
\begin{equation} \label{eq1}
\begin{split}
\nabla& [\mathbf{f(X)} \odot \sigma(\mathbf{g(X)})] = \overbrace{\sigma(\mathbf{g(X)})}^\text{transform gate} \odot \nabla \mathbf{f(X)}   \\ &+ \overbrace{\big( 1 - \sigma(\mathbf{g(X)})\big)}^\text{carry gate} \big( \sigma(\mathbf{g(X)}) \odot \mathbf{f(X)} \big)
\end{split}
\end{equation}{where the $\sigma(.)$ can be seen as the transform gate, while $(1-\sigma(.))$ can be seen as the carry gate. This could be regarded as a form of highway networks.}

% 1. softmax(qK^T)v -> channel-wise | 
% weighted avg -> seq2seq level attn -> moderate the value by control the attention distribution. -> independent of each other -> example two might be both large enough, rather merely under the same sequence-independent distribution.
% multi-head -> multiple semantic subspaces -> can be pruned / not guarantee ->  

\subsection{Variant Gated Connections}
\label{sec:var-sdu}
\begin{figure}[]
\vskip 0mm
\begin{center}
\includegraphics[width=\columnwidth]{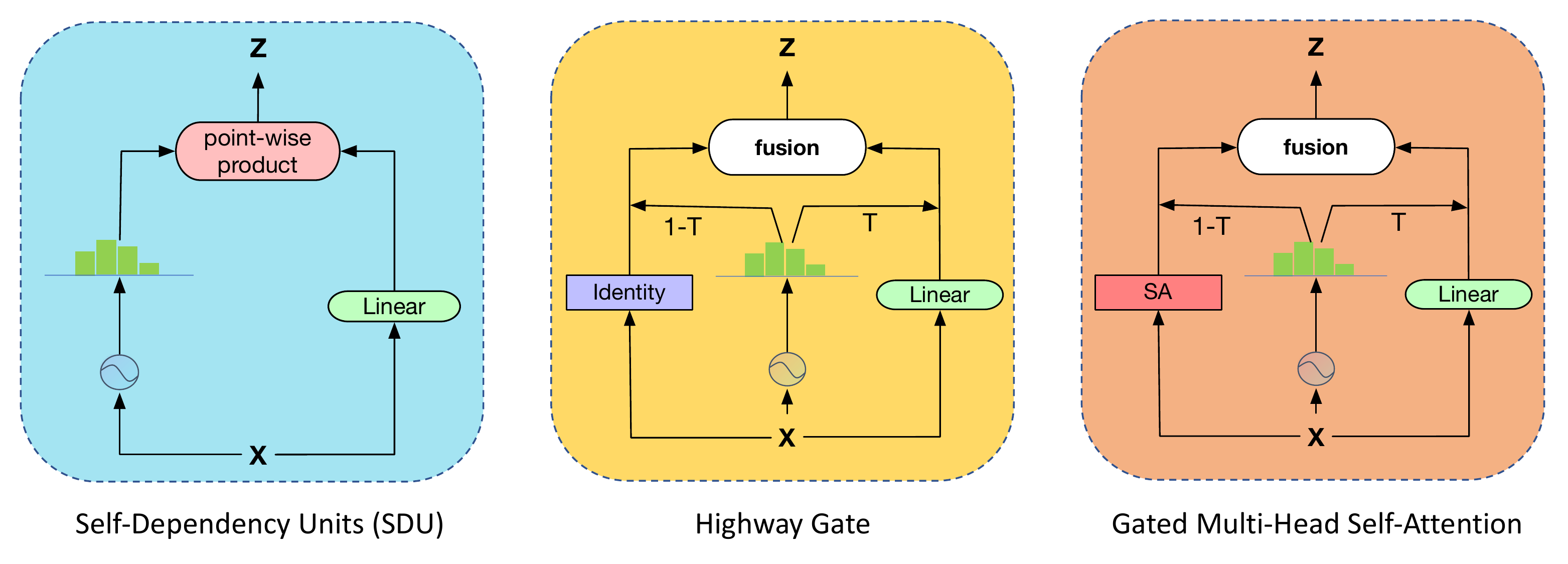}
\vskip -3mm
\caption{SDU variants.}
\label{fig:sdu_variants}
\end{center}
\vskip -5mm
\end{figure}

\paragraph{Highway Gate}
Similar to the highway networks~\citep{srivastava2015highway}, let $T(\mathbf{X})$ signal the \emph{transform} gate and $(1-T(\mathbf{X}))$ be the \emph{carry} gate, we have the highway-network-like structures by regulating the encoding $f(\mathbf{X})$ with \emph{transform} gate and controling $\mathbf{X}$ with \emph{carry} gate. This is quite similar to highway networks:
\begin{align}
\begin{split}\label{eq:hi0}
    T(\mathbf{X}) ={}& \mathbf{\sigma}(\mathbf{X}\mathbf{W_1}+\mathbf{b_1})
\end{split}\\
\begin{split}\label{eq:hi1}
    f(\mathbf{X}) ={}& \mathbf{X}\mathbf{W_2}+\mathbf{b_2}
\end{split}\\
\begin{split}\label{eq:hi2}
    o(\mathbf{X}) ={}& (1- T(\mathbf{X})) \odot \mathbf{X} \\ & + T(\mathbf{X}) \odot f(\mathbf{X})
\end{split}\\
\begin{split}\label{eq:hi3}
    \mathbf{U} ={}& LN \big( o(\mathbf{X}) +\textrm{Att}(\mathbf{Q},\mathbf{K},\mathbf{V}) \big)
\end{split}
\end{align}{where Eq. \ref{eq:hi2} is the element-wise summation of highway networks, $o(\cdot)$ represents the intermediate output (see Fig.~\ref{fig:sdu_variants}).}

% Similarly, the FFN modules have:
% \begin{align}
% \begin{split}
%     T'(\mathbf{u}) ={}& \mathbf{\sigma}(\mathbf{u}\mathbf{W_1}+\mathbf{b_1})
% \end{split}\\
% \begin{split}
%     f(\mathbf{u}) ={}& \mathbf{u}\mathbf{W_2}+\mathbf{b_2}
% \end{split}\\
% \begin{split}
%     o(\mathbf{u}) ={}& (1- T'(\mathbf{u})) \odot \mathbf{u}  + T'(\mathbf{u}) \odot f(\mathbf{u})
% \end{split}\\
% \begin{split}
%     out ={}& LN \big( o(\mathbf{u}) +\textit{Att}(\mathbf{Q},\mathbf{K},\mathbf{V}) \big)
% \end{split}
% \end{align}

\paragraph{Gated MHDPA}
Similar to previous highway gates, we can apply the carry gate and transform gate on the attention and FFN units respectively. Thus we have:
\begin{align}
% \begin{split}
%     T(\mathbf{X}) ={}& \mathbf{\sigma}(\mathbf{X}\mathbf{W_1}+\mathbf{b_1})
% \end{split}\\
% \begin{split}
%     f(\mathbf{X}) ={}& \mathbf{X}\mathbf{W_2}+\mathbf{b_2}
% \end{split}\\
\begin{split}
    o(\mathbf{X}) ={}& (1- T(\mathbf{X})) \odot \textrm{Att}(\mathbf{Q},\mathbf{K},\mathbf{V})\\& + T(\mathbf{X}) \odot  f(\mathbf{X})\label{eq:gSA3}
\end{split}\\
\begin{split}
    \mathbf{U} ={}& LN \big( o(\mathbf{X}) + \mathbf{X} \big)\label{eq:gSA4}
\end{split}
\end{align}{Such gates can be regarded as dynamically adjusting the information flow between the feature-wise representations and SANs (Eq.~\ref{eq:gSA3}).}

% ResNet / Highway $H(x) - x$

% A).  Highway masked gate (and explanation)
% 1. input
% x*sigmoid(x) forget gate -> complicated version of ReLU mask
% 2. forget
%  x* (1-sigmoid(x))
% 3. control gate
% x*tanh(x)
% note: we also try others and find it not as efficient as this

% note: why add? concat and FC is not empirically as good as this

% for shallow networks, better / deep -> overfitting

% B).
% 1. softmax(qK^T)v -> channel-wise | 
% weighted avg -> seq2seq level attn -> moderate the value by control the attention distribution. -> independent of each other -> example two might be both large enough, rather merely under the same sequence-independent distribution.
% multi-head -> multiple semantic subspaces -> can be pruned / not guarantee ->  

% sigmoid(q)V -> lattice-wise. Model, the Bernoulli Distribution.->Self gate controls the in

%  $\tanh(x) = 2 \sigma(2x)-1$
% tanh() -> 1- 2\sigma -> better symmetry character. Both insensitive to too large values.

\section{Experiments and Results}
\label{sec:exp}
We apply the gating mentioned above on Transformer variants described in section~\ref{sec:pre} on LM tasks and respectively make comparisons in terms of both the convergence process and the final performance.
For fairness, we apply SDU components based on the same hyperparameters as the original paper\footnote{Some results of baselines are slightly lower than those reported in original papers using the code obtained from authors but are within the limits of experimental error and variance.}. Our code is available\footnote{\url{https://github.com/cyk1337/Highway-Transformer}}.
% No hyperparameter tuning on gating enhanced models has been conducted in all the experiments to eliminate the impact of variable experimental settings. 
% fairly compare the experimental results under the same experimental settings.
% canonical model architectures in PyTorch.
% Due to the limited computing resources, we mainly experiment on Transformer models with shallow layers and finally have a few trials on the 12-layer Transformer-XL architecture. 

\subsection{vs. Transformer / R-Transformer}
We first evaluate the gating units on the Penn TreeBank (PTB) LM task. The SDU gates are added on Eq.~\ref{eq:att_m}, \ref{eq:fc_m} for each Transformer block. All models in this section are trained on single NVIDIA Titan Xp GPU.

% \begin{figure}[thb]
\begin{figure}[t]
\vskip 0mm
\begin{center}
\includegraphics[width=\columnwidth]{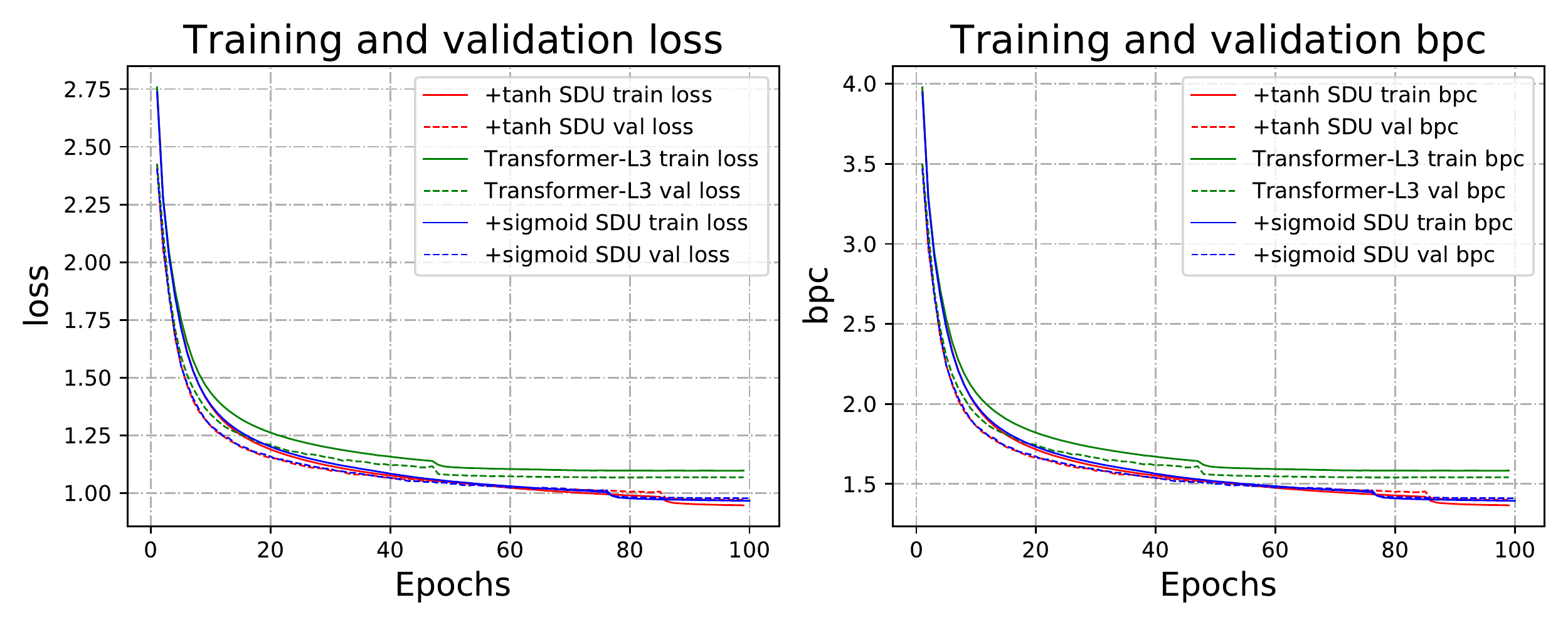}
\vskip -3mm
\caption{The 3-layer \textbf{Transformer}'s curve of training and evaluation performance on character-level PTB LM.}
\label{fig:T-L3-char}
\end{center}
\vskip -7mm
\end{figure}

\subsubsection{Char-level PTB}
\paragraph{Hyperparameter and Training}
The gated components are evaluated on character-level PTB LM tasks (see Appendix~\ref{ap:hp-char-PTB} for hyperparameter settings). The loss and bit per character (bpc) provide the metrics to evaluate the trained models. All models are trained with 100 epochs.

% \begin{figure}[thb]
\begin{figure}[]
\vskip 0mm
\begin{center}
\includegraphics[width=\columnwidth]{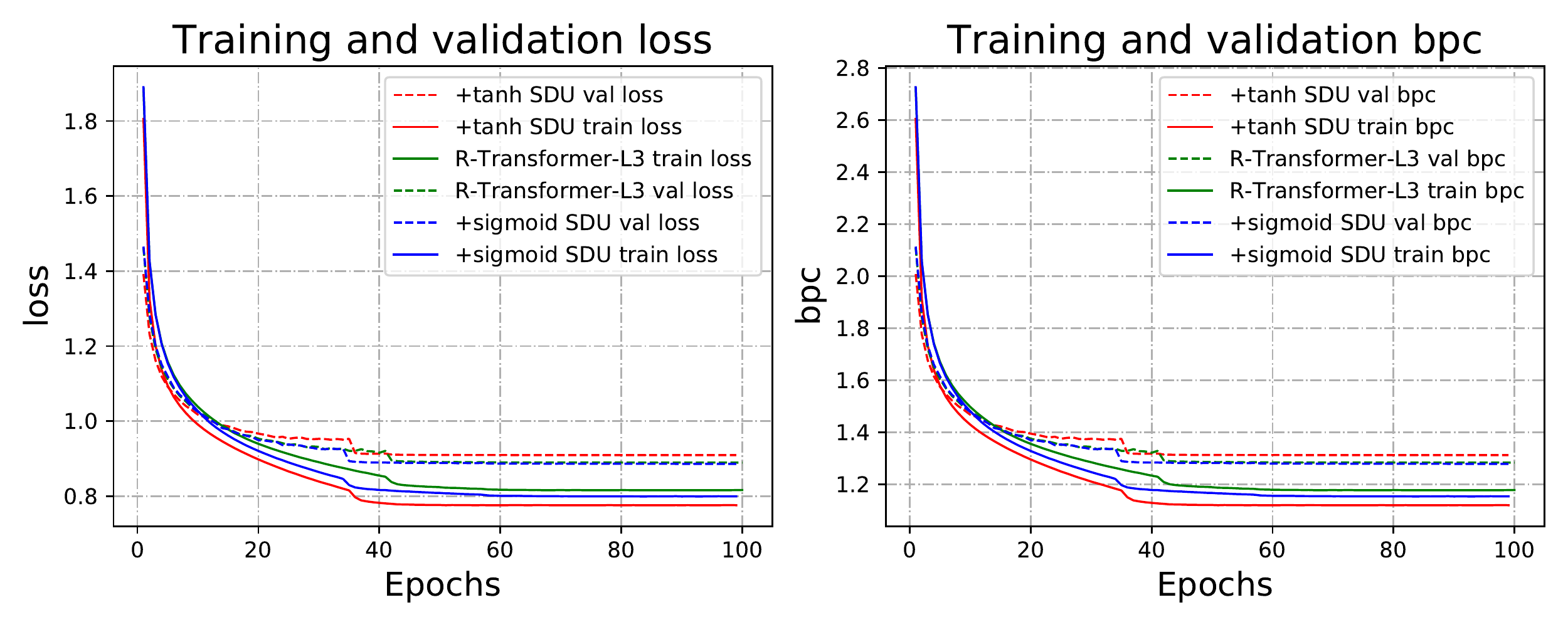}
\vskip -3mm
\caption{The 3-layer \textbf{RT}'s curve of training and evaluation performance on character-leval PTB LM task.}
\label{fig:RT-L3-char}
\end{center}
\vskip -5mm
\end{figure}

\paragraph{Results of Transformer}
As shown in Table~\ref{tab:T-ptb-char}, all the gating-enhanced models conspicuously surpass the performance of the loss and perplexity over the baseline on both training and validating set, revealing the positive influence of self-gating units in supporting Transformer blocks. Furthermore, Fig.~\ref{fig:T-L3-char} presents the beneficial effect of gating units in accelerating the convergence process in both training and evaluation set by a clear margin, validating the accumulative effect that our gating units bring out. In which SDUs with $\tanh$ gates (8.76\% improvement) outperform the counterpart with sigmoid gates (8.2\% improvement) in terms of the final perplexity on the test set.

%  It is observed that kick-points of decline in the loss and bpc curves have a distinct postponement: 3-layer attention models with SDU components reveal a much rapid decrease and 30 to 40 epochs slower to encounter the kickpoints. Meanwhile, SDU with $\tanh$ meets the turning point approximately 10 epochs later than the sigmoid gates and finally reach a better place at the 100-th epoch. 

\begin{table}[!ht]
\scalebox{0.9}{
\begin{tabular}{@{}lllll@{}}
\toprule
model & eval loss & eval ppl & test loss & test ppl \\ \midrule
T-L3 & 1.068 & 1.541 & 1.036 & 1.495\\
+$\sigma$ SDU & 0.9776 & 1.410$\Downarrow$ & 0.950 & 1.371$\Downarrow$ \\
+$\tanh$ SDU & \textbf{0.9714} & \textbf{1.401}$\Downarrow$ & \textbf{0.945} & \textbf{1.364}$\Downarrow$ \\ \bottomrule
\end{tabular}
}
\caption{Performance of 3 Layer \textbf{Transformers} and SDU components on char-level PTB LM task. The best performance is marked bold.}
\label{tab:T-ptb-char} \vskip -5mm
\end{table}

\paragraph{Results of RT}
It can be seen in Fig.~\ref{fig:RT-L3-char} that supplementing SDUs can increase the speed of the convergence process of training and evaluation, strengthening our previous claim. As for the final perplexity on the test set, $\sigma$-gate SDUs could achieve better than baselines while $\tanh$-gate SDUs perform a bit worse, as shown in Table~\ref{tab:RT-ptb-char}. The influence of $\sigma$-gate SDUs might be owing to that $\sigma$ function compresses the input into the dense non-zero ratios within $(0,1)$ and results in stable variation range. In contrast, the zero-centered property and possibly zeroed values of $\tanh$ may cause the corresponding units easier to be trapped into the premature convergence during the training process. Besides, $\sigma$ gates have been empirically proved to be more stable than $\tanh$ gates in the follow-up experiments.

% pushes the representations into possibly zeroed outputs in the range of $(-1,1)$ and may be of more variation.
% TODO
% We attribute this to the vulnerability to the $\tanh$ gate's premature convergence phenomenon during the evaluation process and will discuss this in the following section.

% It is observed in Fig.~\ref{fig:RT-L3-char} that support the evidence from previous experiments that models with gated units all performed a significantly rapid constringency speed in terms of training. Similarly, SDUs make RTs converge nearly 10 epochs faster out of 100 epochs in terms of the dev loss and perplexity.

\begin{table}[!ht]
\scalebox{0.9}{
\begin{tabular}{@{}lllll@{}}
\toprule
model & eval loss & eval ppl & test loss & test ppl \\ \midrule
RT-L3 & 0.8896 & 1.283 & 0.867 & 1.250 \\
+$\tanh$ SDU & 0.9096 & 1.312 & 0.883 & 1.274 \\
+$\sigma$ SDU & \textbf{0.8863} & \textbf{1.279}$\Downarrow$ & \textbf{0.863} & \textbf{1.245}$\Downarrow$ \\ \bottomrule
\end{tabular}}
% \vskip -1mm
\caption{Performance of 3 Layer \textbf{R-Transformers} and SDU components on char-level PTB LM task.}
\label{tab:RT-ptb-char} \vskip -4mm
\end{table}

\subsubsection{Word-level PTB}
\paragraph{Hyperparameter and Training}
We compare the performance between 3-layer Transformer and R-Transformer (RT) with and without SDU gating units. Appendix~\ref{ap:hp-char-PTB} illustrates the hyperparameter setup. All experiments are conducted with 100 epochs, and the loss and perplexity (ppl) values on the development set serve as evaluation metrics.

% \begin{figure}[thb]
\begin{figure}[ht]
\vskip 0mm
\begin{center}
\includegraphics[width=\columnwidth]{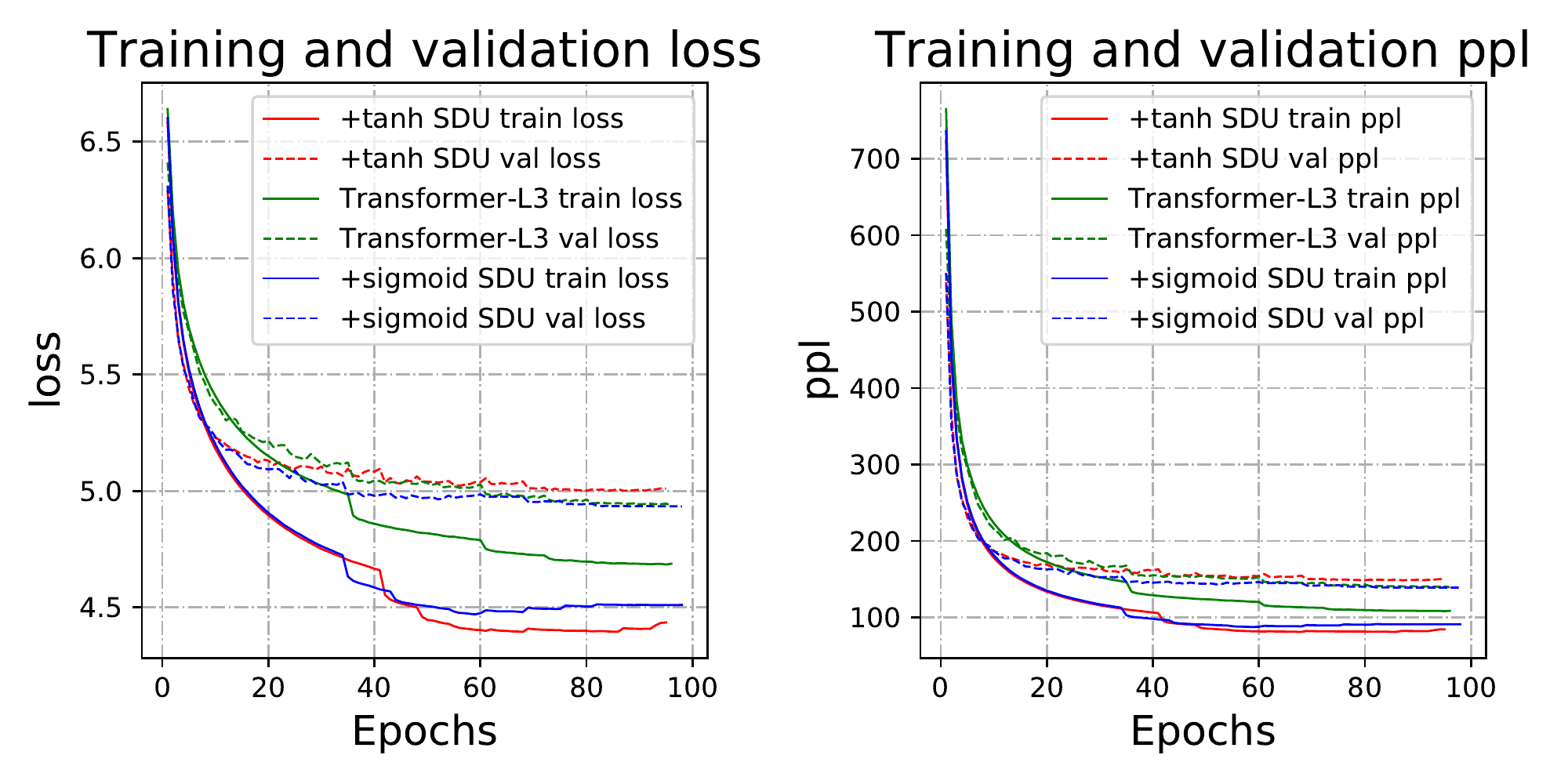}
\vskip -3mm
\caption{Loss and perplexity of 3-layer \textbf{Transformers} on the word-level PTB training and validation set.}
\label{fig:T-L3-word}
\end{center}
\vskip -3mm
\end{figure}

\begin{table}[ht]
\scalebox{0.9}{
\begin{tabular}{@{\extracolsep{\fill}}lllll@{}}
\toprule
model & eval loss & eval ppl & test loss & test ppl \\ \midrule
T-L3 & 4.937 & 139.4 & \textbf{4.87} & 130.43 \\
+$\sigma$ SDU & \textbf{4.934} & \textbf{138.9}$\Downarrow$ & \textbf{4.87} & \textbf{130.30}$\Downarrow$ \\
+$\tanh$ SDU & 5.001 & 148.5 & 4.94 & 139.53 \\ \bottomrule
\end{tabular}
}\vskip -2mm
\caption{Performance of 3-layer basic \textbf{Transformer} (T-L3) and SDU components on word-level PTB LM.}
\label{tab:T-ptb-word} \vskip -4mm
\end{table}

\paragraph{Results of Transformer} Figure~\ref{fig:T-L3-word} shows a noticeable downward trend on the evaluation performance (i.e., the validation loss and perplexity) of the attention model with $\tanh$ and sigmoid functions over the beginning 30 epochs, again indicating the convergence acceleration effect of our gated units. Also, $\sigma$-gate enhanced models outmatches the baseline on the test perplexity, but models with $\tanh$ gates reach into a plateau untimely. As for the training curves, Transformers with SDUs have seen a remarkably sharper fall in comparison with the baseline model over all the training period.

\begin{figure}[ht]
\vskip 0mm
\begin{center}
\includegraphics[width=\columnwidth]{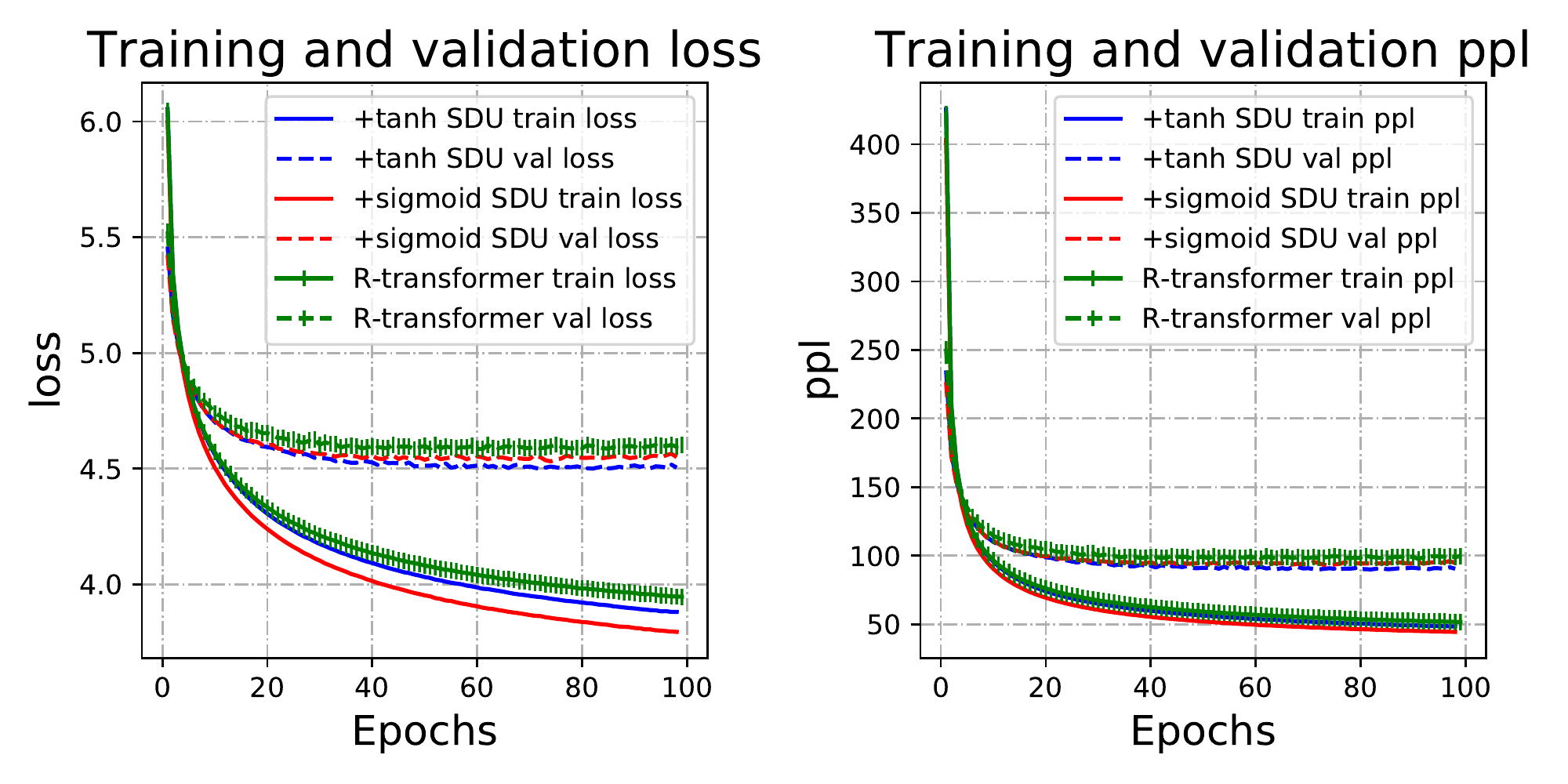}
\vskip -3mm
\caption{Loss and perplexity of 3-layer \textbf{RT} on the word-level PTB training and validation sets.}
\label{fig:RT-L3-word}
\end{center}
\vskip -8mm
\end{figure}

\paragraph{Results of RT} As in Fig.~\ref{fig:RT-L3-word} and Table~\ref{tab:R-ptb-word}, models with SDUs entirely surpass the performance of the baseline involving both the convergence speed and perplexity on the test set. Similar to the word-level R-Transformer, $\tanh$-gate SDUs behave a bit better than the counterpart with sigmoid gates, both showing stable curvatures of convergence.

\begin{table}[ht]
\scalebox{0.9}{
\begin{tabular}{@{}lllll@{}}
\toprule
model & eval loss & eval ppl & test loss & test ppl \\ \midrule
RT-L3 & 4.58 & 97.63 & 4.53 & 92.31 \\
+$\sigma$ SDU & 4.53 & 92.91$\Downarrow$ & 4.48 & 87.88$\Downarrow$ \\ 
+$\tanh$ SDU & \textbf{4.50} & \textbf{89.97}$\Downarrow$ & \textbf{4.44} & \textbf{84.92}$\Downarrow$ \\ 
\bottomrule
\end{tabular}
}\vskip -2mm
\caption{The performance of 3-layer R-Transformers (RT-L3) and SDU components on word-level PTB LM.}
\label{tab:R-ptb-word}
\vskip -5mm
\end{table}

% \begin{figure}[thb]

\subsection{Sub-total}
To sum up, gating units have empirically expedited the convergence of Transformer blocks due to the enrichment of self-regulated features with skip-connections. It can be seen that $\sigma$-gate presents the stability to bear a hand to reach the plateau without hurting the test performance, but $\tanh$-gate seems to be task- and data-dependent and could be better than $\sigma$-gate SDUs in some circumstances. We can see that our proposed gated units are complementary to the recurrent connections in RNNs and can boost the performance based on \emph{localRNN}-encoded representations.

In the following experiment, we check whether it is necessary to apply gates on all the layers and probe the effect of SDU variants (i.e., ``\emph{highway gate}'' and ``\emph{gate MHDPA}''). Due to the small size of PTB, we experiment on a larger LM dataset \emph{enwik8} and adopt the impressive Transformer-XL, one of the vital variant structures used in XLNet~\citep{yang2019xlnet}.

\subsection{vs. Transformer-XL}
% We evaluate proposed gates on Transformer-XL, on \emph{enwik8} dataset. 
% We firstly make experiments on 6-layer shallow networks and then have plenty of explorations to test the performance of 12-layer deep-stacked networks.

% TODO

\paragraph{Hyperparameter}
See Appendix~\ref{ap:hp-xl} for detailed hyperparameter settings.

% Models of 6 and 12 layers are trained on 4 x 2080Ti GPUs in parallel, of each spending $\approx$18 and $>$100 hours for 40k and 400k steps, respectively. 
% Due to the limited computing resources, we cannot afford the extensive grid-search-like attempts. Thus we mainly focus on probing the effect of augmenting SDU gates on layers of different depth.

\subsubsection{Results of 6-layer Transformer-XL}
\begin{figure}[!ht]
% \vskip 0mm
\begin{center}
\includegraphics[width=\columnwidth]{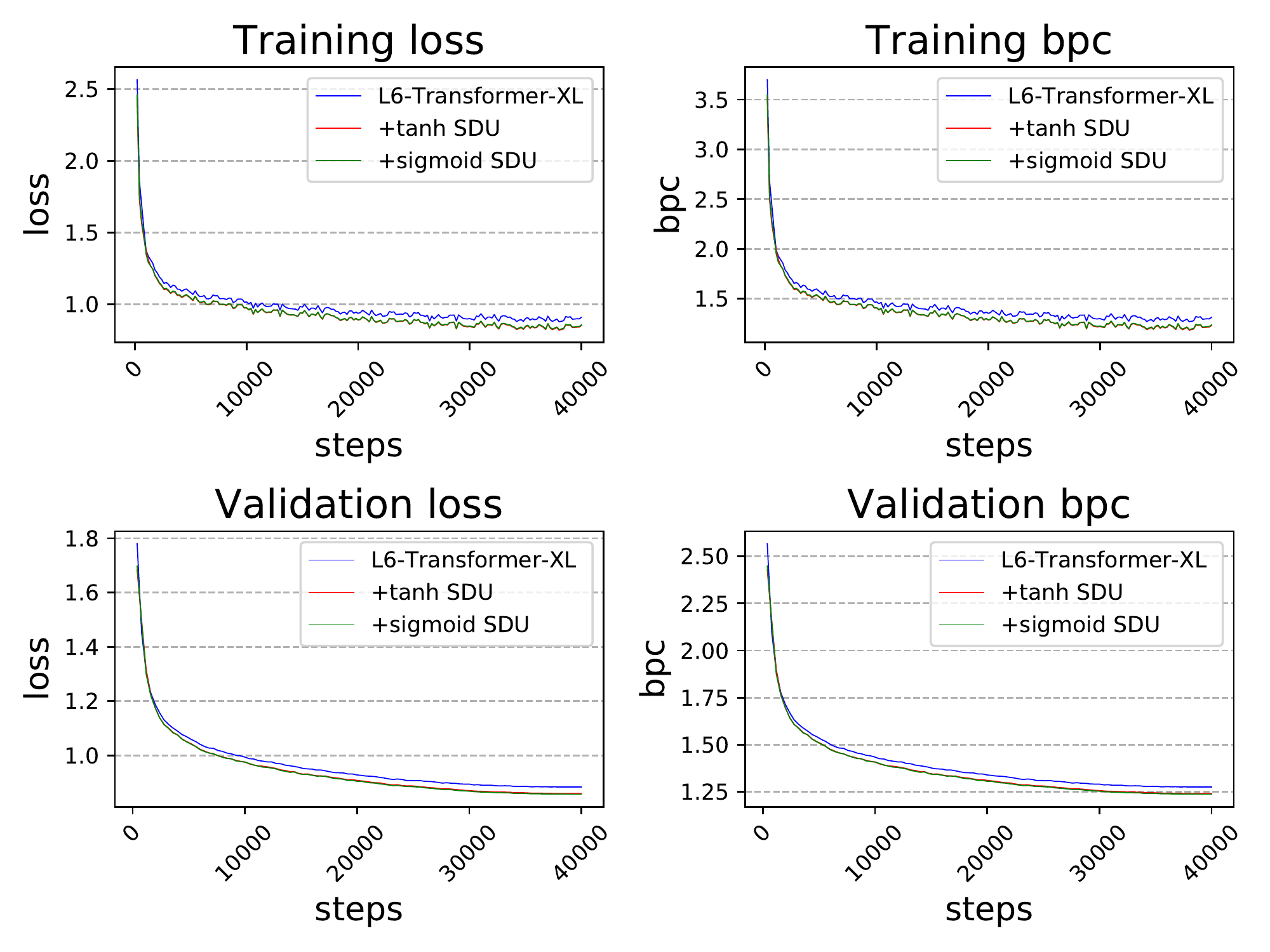}
\vskip -3mm
\caption{The comparison between 6-layer Transformer-XL with adding different SDUs.}
\label{fig:xl-L6-best}
\end{center}
\vskip -4mm
\end{figure}

It is noticeable that Transformer-XL models with different gating variants all outperform the baseline with different margins in terms of both performance and convergence speed, as shown in Table~\ref{tab:xl-L6}. Fig.~\ref{fig:xl-L6-best} shows that SDUs benefit the convergence and validation performance compared with baselines. Among which $\sigma$-gate SDUs ranked top by achieving 3.1\% improvement of bpc on the dev set, followed by \emph{gates with tanh}, \emph{gated MHDPA}, \emph{highway gate} with 2.7\%, 1.8\%, 1.7\% advance respectively. We attribute such improvements to the augmented refined representations learned by our gated units, preventing the basic self-attention blocks from purely considering the contextual dependency. It is also illustrated that SDUs do not conflict with recurrence mechanisms in Transformer-XL.

\begin{table}[]
\scalebox{0.75}{
\begin{tabular}{@{}lllll@{}}
\toprule
\textbf{model} & \textbf{eval loss} & \textbf{eval bpc} & \textbf{test loss} & \textbf{test bpc} \\ \midrule
L6-XL & 0.8843 & 1.276 & 0.86 & 1.24339 \\\cmidrule{1-1}
+$\tanh$ SDU & 0.8602 & 1.241$\Downarrow$ & \textbf{0.84} & 1.21424$\Downarrow$ \\
+$\sigma$ SDU & \textbf{0.8577} & \textbf{1.237}$\Downarrow$ & \textbf{0.84} & \textbf{1.21123}$\Downarrow$ \\ 
+highway gate & 0.8692 & 1.254$\Downarrow$ & 0.85 & 1.22177$\Downarrow$ \\
+gated MHDPA & 0.8682 & 1.253$\Downarrow$ & 0.85 & 1.22398$\Downarrow$\\ \midrule
\multicolumn{5}{c}{\textbf{Ablation study}} \\ \midrule
+$\tanh$  L1-6\textbackslash FFN & 0.8720 & 1.258$\Downarrow$ & 0.85 & 1.22866$\Downarrow$ \\
+$\tanh$ L1-3 & \textbf{0.8660} & \textbf{1.249}$\Downarrow$ & \textbf{0.85} & \textbf{1.22039}$\Downarrow$ \\
+$\tanh$ L3-6 & 0.8852 & 1.277$\Downarrow$ & 0.86 & 1.24420$\Downarrow$ \\\cmidrule{1-1}
+$\sigma$  L1-6\textbackslash FFN & 0.8752 & 1.263$\Downarrow$ & 0.85 & 1.23332$\Downarrow$ \\
+$\sigma$ L1-3 & 0.8792 & 1.268$\Downarrow$ & 0.86 & 1.23589$\Downarrow$ \\ 
+$\sigma$ L3-6 & 0.8843 & 1.276$\Downarrow$ & 0.86 & 1.24261$\Downarrow$ \\
\bottomrule
\end{tabular}
}\vskip -2mm
\caption{Results of \textbf{6-layer Transformer-XL} (L6-XL) and augmented SDUs with different settings. ``\textit{+$\sigma$'' L1-6\textbackslash FFN} represents adding $\sigma$-SDUs on MHDPAs of 1-st to 6-th layers but not on FFN sublayers.}
\label{tab:xl-L6} \vskip -6mm
\end{table}

\subsubsection{Ablation Study}
\paragraph{6-layer Transformer-XL}
To probe whether it is required to augment SDUs on each Transformer layer, we supplement gates on layer 1-3, layer 3-6, and layer 1-6 but removing gates on FFN components (denoted ``\textbackslash FFN'') as in Table~\ref{tab:xl-L6} (see Fig.~\ref{ap-fig:xl-L6} in Appendix~\ref{ap-sec:exp-xl-L6} for detailed convergence curvatures). We find that supplementing $\tanh$-gates on the bottom three layers contribute most to the overall performance while $\tanh$-gates on the top three layers could hinder the test set performance. Low-level Transformer blocks can capture the information from localness while top layers usually focus on the global long-range dependency~\citep{yang2018modeling}. Thus gates on bottom layers could aid in learning syntax and superficial representations to some extent. It also indicates that our gates may be beneficial for the encoding of low-level fine-granularity representations rather than semantic meaning regulation on high-level layers. 
% We also find that removing the gates on FFN in each layer could also help deep models to generalize.

 % Further, we investigate the effectiveness of gating units on different levels of the stacked model, including removing SDUs on $FFN$s and only keeping the MHDPA parts (e.g. ``\emph{tanh \textbackslash FFN}'' in the Table~\ref{tab:xl-L6}) and SDUs on the different depth of layers (e.g., ``\emph{$\sigma$ L3-6}'' represents adding sigmoid functions on the 3-th to 6-th layers). The results (see Table~\ref{tab:xl-L6} and Fig.~\ref{fig:xl-L6}) demonstrate that gating units on shallow layers (i.e., L1-3 in the table) hold the promise of boosting the performance, while replenishing SDU gates on the tail layers do not show a noticeable effect. It can be inferred that gating units on FFN components count for the overall improvements because removing the gating units of $\tanh$ and sigmoid gates on $FFN$s results in the drop of 49.4\% and 68.7\% w.r.t the performance improvements.

\paragraph{12-layer Transformer-XL}
Previous experiments are all conducted on shallow models and illustrate the positive effects. To investigate the performance on deep stacked models, we further extend our trials to 12-layer Transformer-XL. All hyperparameters are the same as 6-layer Transformer-XL, as shown in Appendix~\ref{ap:hp-xl}. Each model is trained 400k steps for more than 100 hours on 4 x GeForce 2080Ti GPUs in parallel.

\begin{figure}[!ht]
% \vskip 0mm
\begin{center}
\includegraphics[width=\columnwidth]{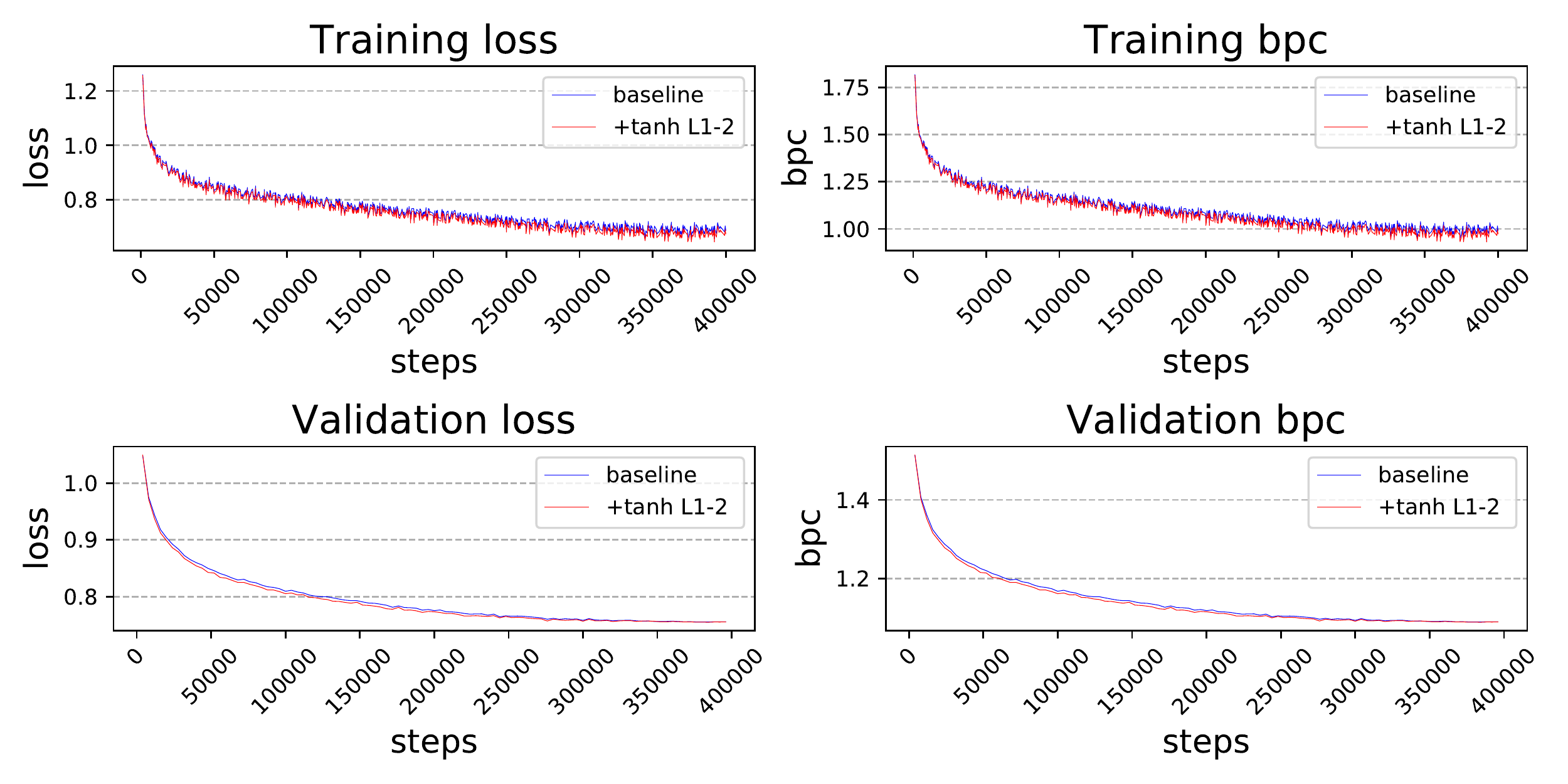}
\vskip -3mm
\caption{The comparison between \textbf{12-layer Transformer-XL} with and without $\tanh$ gated units on bottom two layers.}
\label{fig:xl-L12-best}
\end{center}
\vskip -4mm
\end{figure} 

\begin{table}[]
\scalebox{0.75}{
\begin{tabular}{@{}lllll@{}}
\toprule
\textbf{model} & \textbf{eval loss} & \textbf{eval bpc} & \textbf{test loss} & \textbf{test bpc} \\ \midrule
L12-XL & 0.7554 & 1.090 & 0.74 & 1.07160 \\\midrule
\multicolumn{5}{c}{\textbf{Ablation study}} \\ \midrule
+$\tanh$ L1-12 & 0.7919 & 1.143 & 0.78 & 1.12797 \\
+$\tanh$ L1-6 & 0.7623 & 1.100 & 0.75 & 1.08234 \\
+$\tanh$ L1-3 & 0.7558 & 1.090 & 0.74 & 1.07140$\Downarrow$ \\
+$\tanh$ L1-2 & 0.7548 & \textbf{1.089}$\Downarrow$ &0.74 & \textbf{1.06904}$\Downarrow$ \\
+$\tanh$ L1 & 0.7549 & \textbf{1.089}$\Downarrow$ & 0.74 & 1.06960$\Downarrow$ \\
+$\tanh$ L6-12 & 0.7572 & 1.092 & 0.74 & 1.07313 \\
+$\tanh$ \textbackslash $FFN$ & 0.7734 & 1.116 & 0.76 & 1.09920 \\
\cmidrule{1-1}
+$\sigma$ L1-12 & 0.7752 & 1.118 & 0.77 & 1.10462 \\
+$\sigma$ L1-6 & 0.7635 &1.101 & 0.75 & 1.08283 \\ 
+$\sigma$ L1-3 & 0.7580 & 1.094 & 0.74 & 1.07383\\ 
+$\sigma$ L1-2 & 0.7552 & 1.090 & 0.74 & 1.07148$\Downarrow$ \\ 
+$\sigma$ L1 & 0.7557 & 1.090 & 0.74 & 1.07157$\Downarrow$ \\ 
+$\sigma$ L6-12 & 0.7585 & 1.094 & 0.75 & 1.07607 \\
+$\sigma$ \textbackslash $FFN$ & 0.7647 & 1.103 & 0.75 & 1.08652 \\
\midrule
+highway gate & 0.7784 & 1.120 & 0.77 & 1.10922 \\
+gated MHDPA & 0.7741 & 1.117 & 0.76 & 1.10292 \\ 
\bottomrule
\end{tabular}
}
% \vskip -2mm
\caption{Final results of \textbf{12-layer Transformer-XL} (XL-L12) and augmented SDUs with different settings.}
\label{tab:xl-L12} 
\vskip -3mm
\end{table}

The experimental results illustrate that SDU components have contributed to expediting the convergence during training (see Fig.~\ref{ap-fig:xl-L12-tanh} and~\ref{ap-fig:xl-L12-sigmoid} in Appendix~\ref{ap-sec:exp-xl-L12} for details). But supplementing gated units on each Transformer block could encounter the premature convergence phenomenon. It is also observed that adding the bottom few layers with gated units could strengthen the convergence process without impeding the final performance, as shown in Table~\ref{tab:xl-L12}. It is observed from Fig.~\ref{fig:xl-L12-best} that $\tanh$-gates on the bottom two layers promote the convergence process and further improve the bpc performance on the dev and test set.

Interestingly, the performance does not follow a positive correlation with the increase of gated layer numbers. We can see that enriching the bottom 2 layers with $\tanh$ and $\sigma$ gated functions (denoted ``+tanh L1-2'' and ``+$\sigma$ L1-2'' in Table~\ref{tab:xl-L12}) could impressively benefit for the convergence on both training and evaluation process and even marginally increase the final test bpc (see Fig.~\ref{ap-fig:xl-L12-tanh} and Fig.~\ref{ap-fig:xl-L12-sigmoid} in Appendix~\ref{ap-sec:exp-xl-L12} for details). Therefore, the lower layers benefit more from our proposed gated units than higher layers, again illustrating that SDUs could enhance feature-wise information on shallow layers of deep-stacked Transformer components.

% Deep stacked Transformer with either $\sigma$ or $\tanh$ are both liable to the premature convergence in contrast to 12-layer baseline models, as in Fig~\ref{tab:xl-L12}. 
% On the one hand, removing SDU on FFN components on each layer has a relatively better generalization and convergence curve. On the other hand, merely supplementing $\tanh$ gates on the next few layers (e.g., L1-6) could relieve this situation by boosting the decline slope and advance the baseline performance during training. 
% A possible explanation for this might be that SDU enhances the optimization step with fine granularity, which could lead to better fine-tuning on the training set but also easy to be trapped in the local minimum and therefore cause the relatively poor generalization. Analogously, the gradient descent optimizer with refined strides might encounter suboptimal traps while that with more giant steps could learn not quite accurately but have a fair generalization.

% These results suggest that gating units on the top few layers in deep models could speed up the convergence during training without retarding the test performance. For instance, the ``$\tanh$-L3'' setting performs similarly to ``\emph{tanh L1-6}'' over the period at initial stages, but the latter converge in advance. 
% It drops steadily in the following training process and finally matches or slightly outperform baselines. In contrast, the ``\emph{tanh L1-6}'' plateaued after the 220k-th step.

\subsection{Gating Mechanism Analysis}
It can be concluded that gating units could boost the convergence, especially on low-level layers. Enhancing the bottom layers of deep-stacked models may result in faster convergence of optimization. This may be owing to that SDU gates can enrich the original representations with adaptive self-dependency encodings. The final hidden state can be regarded as a revised representation that incorporating additional self-attentive features.

Meanwhile, we find that supplementing SDU gates does not increase much of the time cost in comparison with baselines. Instead, the total running time of each experimental setting is quite similar. We summarize the training time costs of 6-layer Transformer-XL as table~.\ref{tab:time}.

\begin{table}[thb]
\centering
\resizebox{.6\linewidth}{!}{%
\begin{tabular}{@{}lc@{}}
\toprule
\textbf{model} & \textbf{time cost (hour)} \\ \midrule
xl-L6 & 21.16 \\
+$\tanh$ SDU & 21.45 \\
+$\sigma$ SDU & 21.87 \\
+ highway gate & 21.93 \\
+gated MHDPA & 21.10 \\ \bottomrule
\end{tabular}%
}
\caption{Summary of training time costs of 6-layer Transformer-XL.}
\label{tab:time}
\vskip -3mm
\end{table}

It is argued that low-level transformers learn the local-region information while high-level layers pay more attention to global dependencies~\citep{yang2018modeling}. Our experimental results could verify that gated representation on bottom layers can strengthen the performance by introducing additional gated encodings on localness.

Further, the visualization of learned gate bias parameters of 6-layer and 12-layer models, as shown in Fig.~\ref{ap-fig:heat} in Appendix~\ref{ap-sec:heat}, presenting the layer separation with the increase of layer depth. It has seamlessly verified our previous hypothesis that SDU on shallow layers could promote the learning process and attend to different information with top layers. The scatter plot of Fig.~\ref{ap-fig:cluster-xl} in Appendix~\ref{ap-sec:scatter} indicates that gates on different sublayers learn from different aspects in the identical representation space.

SDUs calculate the output by regulating the information flow of inputs conditioned on themselves. Given the hidden dimension of $d$, the additional cost of trainable parameters on each SDU unit in our experiments is $O(2d(d+1))$. Meanwhile, convolutions along the sequence direction can substitute fully-connected feedforward SDU to curtail the extra parameter cost. Such gating units equip good scalability to attach to different Transformer structures with only minor modification of implementation.

The gradient of our SDU components is:
\begin{align}
\nabla [\mathbf{f(x)} \odot \Psi(\mathbf{g(x)})] &= \nabla \mathbf{f(x)} \odot \Psi(\mathbf{g(x)}) \\ &+ \mathbf{f(x)} \odot \nabla \Psi(\mathbf{g(x)})
\end{align}{where $\mathbf{f,g}$ are linear projections and $\Psi$ takes $\tanh$ or $\sigma$ function. The addition operation of two terms provides an unimpeded information flow, which can be regarded as a multiplicative skip connection~\citep{dauphin2017language} while the second term is usually vanishing due to the derivative of the gating function $\mathbf{\Psi}$. Based on the experimental results, we hypothesize that it could accelerate the optimization process to move towards a local minimum.}

% TODO

% \subsection{Analysis}
% \label{subsec:ana}
% C. analysis
% 1. multi-layer (graph / analysis)
%  L1/3/5/7 (+ tanh v.s. std) plot graph (dash/line)

% 2. multi-head (graph / analysis / who / what / how / why)
% H1/2/4/8  (+tanh v.s std) plot graph
% H=8 performs best, indicating adding such highway flow does not impede the original performance.

% complexity + analysis  ?/? of increment

% self-adaptive / self-dependency

\section{Related Work}
In recent years, there have been plenty of works adopting gating units into CNNs to help learn sequential information. \citet{dauphin2017language} proposed stacked gated CNNs by incorporating GLUs into the 1-dimensional convolution operation, achieving the competitive results in comparison to recurrent models on LM tasks. Based on this, \citet{gehring2017convolutional} augmented the attention mechanism together with GLUs on the convolutional structures, also surpassing the deep LSTMs on NMT tasks. Recently, dynamic convolutions were used to replace MHDPA components in Transformers entirely and also get the impressive results on the WMT-14 dataset~\citep{wu2019pay}.

Amounts of works employed gating mechanisms to modulate self-attention sublayers. Gated-Attention Reader~\citep{dhingra2016gated} introduced gated attention by computing gates on the query encoding to interact with document representations for reading comprehension. \citet{zhang2018accelerating} replaced the first layer of Transformer decoding stacks with an average attention layer by computing forget gates using averaged preceding contextual encodings to regulate the current state information. Distance-based SAN~\cite{im2017distance} and DiSAN~\cite{shen2018disan} put a fusion gate to aggregate the representations after the multi-dimensional self-attention block for natural language inference. \citet{lai2019gated} proposed a gated self-attention memory network with aggregated interactions between input sequences and context vectors for answer selection of question answering. 

Notably, our SDU bears a resemblance to the activation Swish~\cite{ramachandran2017searching} in terms of the equation format. Both of them use the sigmoidal function and self-gating mechanism. However, Swish controls the input gated on itself in a tandem way while the proposed SDU applies the gate after a linear projection and performs using a shunt connection in Transformer stacks.

% However, few previous works focus on incorporating gating mechanisms into SAN blocks. In this paper, we explore the influence of injecting gating units into the Transformer and its variant blocks (as described in section~\ref{sec:pre}) on LM tasks.

\section{Conclusion and Future Work}
Gating-enhanced architecture enjoys both the advantage of MHDPA and self-regulated gating mechanism, allowing for the \emph{pseudo}-highway information flow for better convergence by elastically introducing a few trainable parameters. It outperforms or matches the performance of common Transformer variants without hyperparameter tuning. It is empirically proved that self-gating units on shallow layers could provide more internal representations of importance and significantly benefit for convergence. This also supports the argument that different levels of Transformer components attend to different semantic aspects while lower levels pay more attention to local regions. In the future, it is necessary to interpret the semantics that Transformer layers in different depths can convey, which is beneficial for the computing-efficiency.

% Refer to Appendices~\ref{sec:appendix} and \ref{sec:supplemental} for further information. 

% \section*{Acknowledgments}
% The acknowledgments should go immediately before the references. Do not number the acknowledgments section.
% Do not include this section when submitting your paper for review.

% \newpage

\bibliography{anthology,acl2020}
\bibliographystyle{acl_natbib}

% \clearpage
\newpage

\appendix

\section{Experimental Setup Details}
\subsection{Hyperparameter Settings for RT on Char-level PTB}
\label{ap:hp-char-PTB}
For RT on char-level PTB, we adopt the batch size of 16, gradient clipping with maximum L2 norm of 0.15, layer number of 3, hidden dimension of 512, the sequence length of 400 in char-level, the dropout rate for sublayer connection of 0.15, 8 heads for MHDPA, the initial learning rate of 2, SGD optimizer with linear decay, layer number of 3 in both Transformer and RT models. Weights are initialized with uniform distribution $\mathbf{w} \sim U(-0.1,0.1)$ and biases are all initialized as 0s. The size of GRU cells in \emph{localRNNs} is set to 7 in RT. 

\subsection{Hyperparameter Settings for RT on Word-level PTB}
\label{ap:hp-word-PTB}
 We use the dropout rates of 0.35 and 0.15 for sublayer connections and word embeddings, the initial learning rate of 2, gradient clipping with the maximum L2 norm of 0.35, the hidden dimension of 128, 8-head attention, sequence length of 80 in both Transformer and RT. The weights are initialized with uniform distribution $U(-0.01,0.01)$, and the biases are constant 0s. The optimizer is stochastic gradient descent (SGD) with annealed decay. The \emph{localRNN} context size for  LSTM cells is set to 9 in RT.

\subsection{Hyperparameter Settings for Transformer-XL on enwik8}
\label{ap:hp-xl}
We use layer number of 6, 8 heads for MHDPA with hidden size 64 for each head, hidden size of 2,048 in FFN components, the dropout rate of 0.1 in FFN, embedding size of 512, learning rate 0.00025, memory length of 512, batch size of 22, Adam optimizer without the warm-up strategy. We initialize weights under the Gaussian  $\mathcal{N}(0,1)$ and biases as 0s.

\section{Experimental Results of 6-layer Transformer-XL}
\label{ap-sec:exp-xl-L6}
\begin{figure}[!ht]
% \vskip 0mm
\begin{center}
\includegraphics[width=\columnwidth]{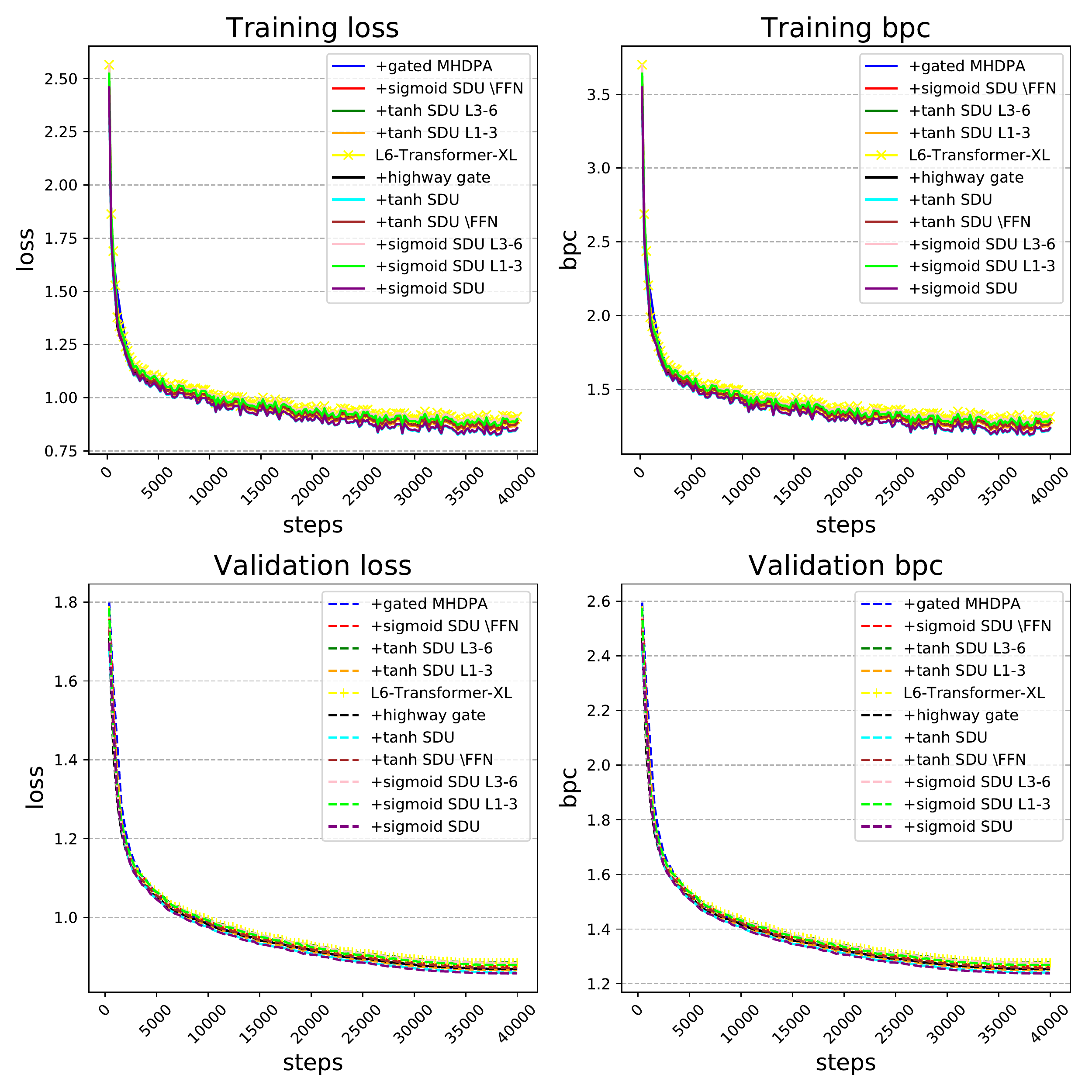}
\vskip -3mm
\caption{The performance of \textbf{6-layer Transformer-XL} experiments with various settings of gated units.}
\label{ap-fig:xl-L6}
\end{center}
\vskip -4mm
\end{figure}

% \subsection{6-layer Transformer-xl}
Fig~\ref{ap-fig:xl-L6} displays all the experimental curvatures with different SDU settings on 6-layer Transformer-XL.
% \begin{figure}[thb]

\section{Experimental Results of 12-layer Transformer-XL}
\label{ap-sec:exp-xl-L12}

% \subsection{12-layer Transformer-XL results}
% \label{ap-sec:xl-12}
% % \begin{figure}[thb]
% \begin{figure}[]
% \vskip 0mm
% \begin{center}
% \includegraphics[width=\columnwidth]{fig/xl-L12.pdf}
% \vskip -3mm
% \caption{The performance of \textbf{12-layer Transformer-XL} experiments with various settings of gated units.}
% \label{fig:xl-L12}
% \end{center}
% \vskip -7mm
% \end{figure} 

\subsection{Transformer-XL v.s. +tanh Gates}
\label{ap-sec:xl-tanh-L12}
\begin{figure}[!ht]
\vskip 0mm
\begin{center}
\includegraphics[width=\columnwidth]{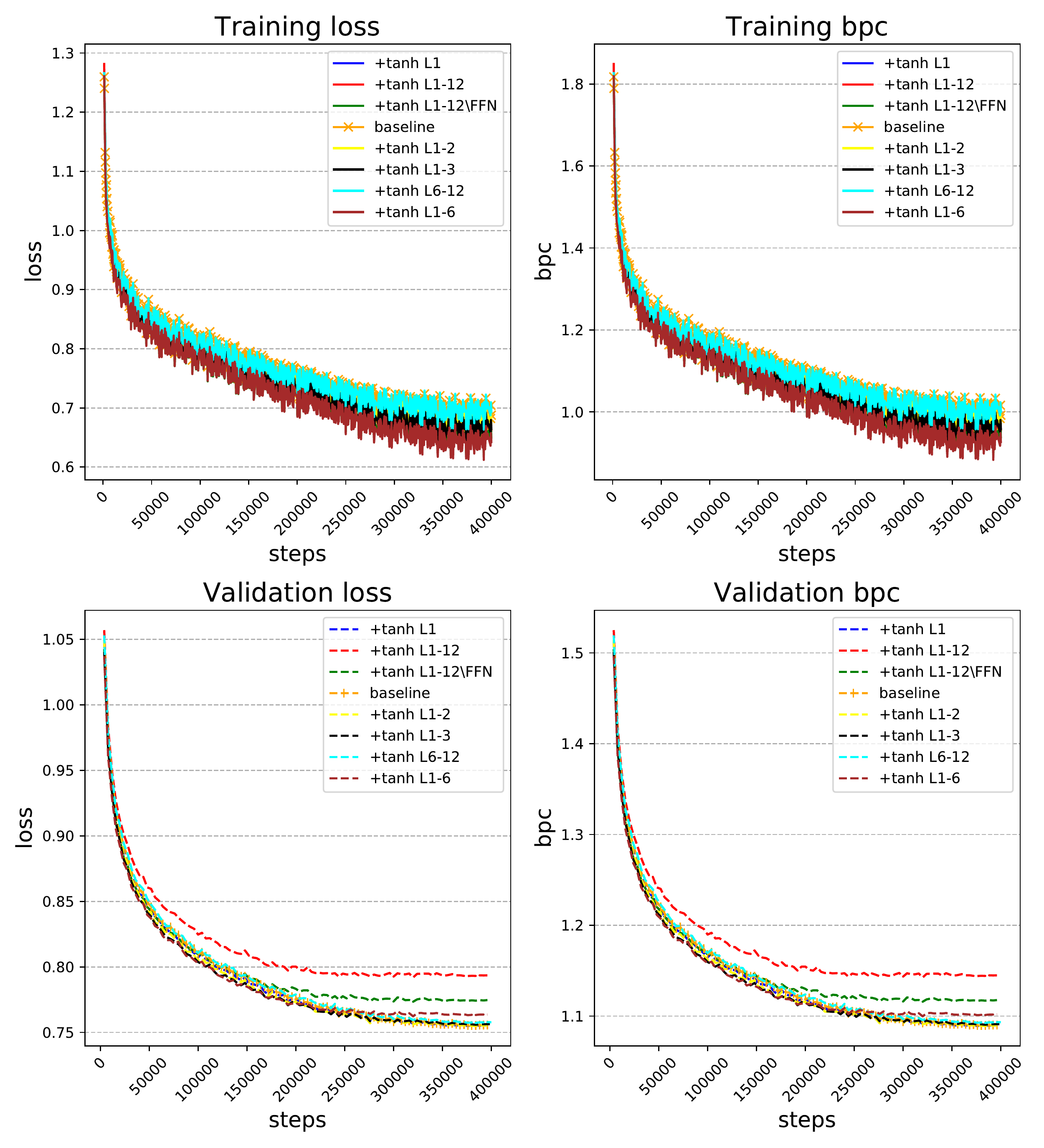}
\vskip -3mm
\caption{The performance of \textbf{12-layer Transformer-XL} experiments augmenting \textbf{$\tanh$} gated units.}
\label{ap-fig:xl-L12-tanh}
\end{center}
\vskip -6mm
\end{figure}

Fig.~\ref{ap-fig:xl-L12-tanh} shows the curve of $\tanh$-gate enhanced Transformer-XL during the training and evaluation process. Adding $\tanh$-gates on the first few layers greatly boost the convergence performance in both the training and evaluation process. Among which ``+tanh L1-2'' presents a rapid convergence trend and marginally outperforms the baseline performance.

% \newpage

\subsection{Transformer-XL v.s. +sigmoid Gates}
\label{ap-sec:xl-sig-xl}

\begin{figure}[!ht]
% \vskip 0mm
\begin{center}
\includegraphics[width=\columnwidth]{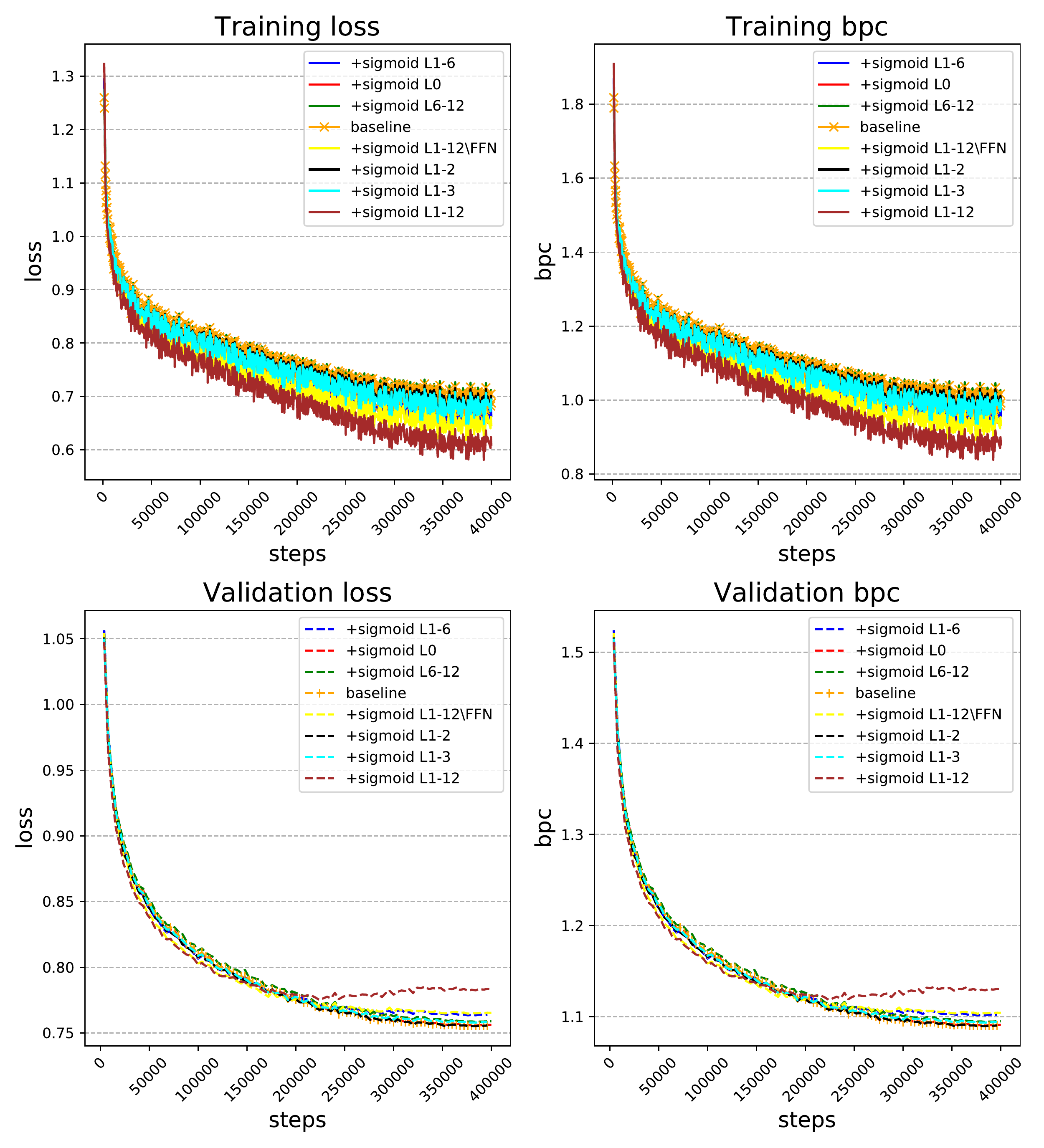}
\vskip -3mm
\caption{The performance of \textbf{12-layer Transformer-XL} experiments augmenting \textbf{$\sigma$} gated units.}
\label{ap-fig:xl-L12-sigmoid}
\end{center}
\vskip -6mm
\end{figure}

Fig.~\ref{ap-fig:xl-L12-sigmoid} illustrates the performance of Transformer-XL augmented with $\sigma$ gates. The sigmoid-gated Transformer-XL has showed a similar trend as $\tanh$ gates in Fig.~\ref{ap-fig:xl-L12-tanh}. 

\section{Plot of Gate Biases of Transformer-XL}
\subsection{Heatmap Visualization}
\label{ap-sec:heat}
\begin{figure*}[t!] % "[t!]" placement specifier just for this example
\begin{subfigure}{0.48\textwidth}
\includegraphics[width=\linewidth]{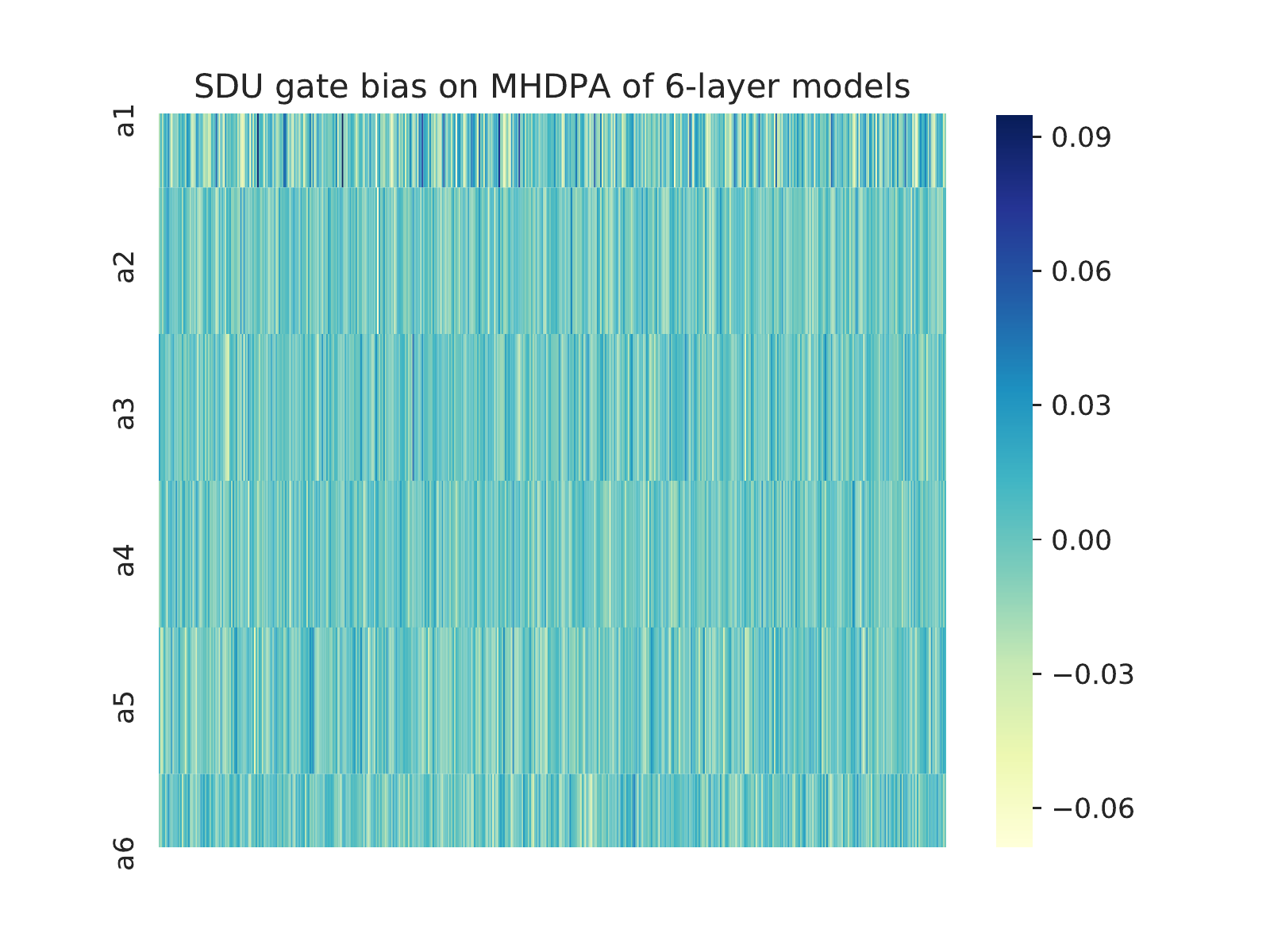}
\caption{Plot of gate biases on MHDPA of 6-layer models.} \label{ap-fig:L6-heat-b-sa}
\end{subfigure}\hspace*{\fill}
\begin{subfigure}{0.48\textwidth}
\includegraphics[width=\linewidth]{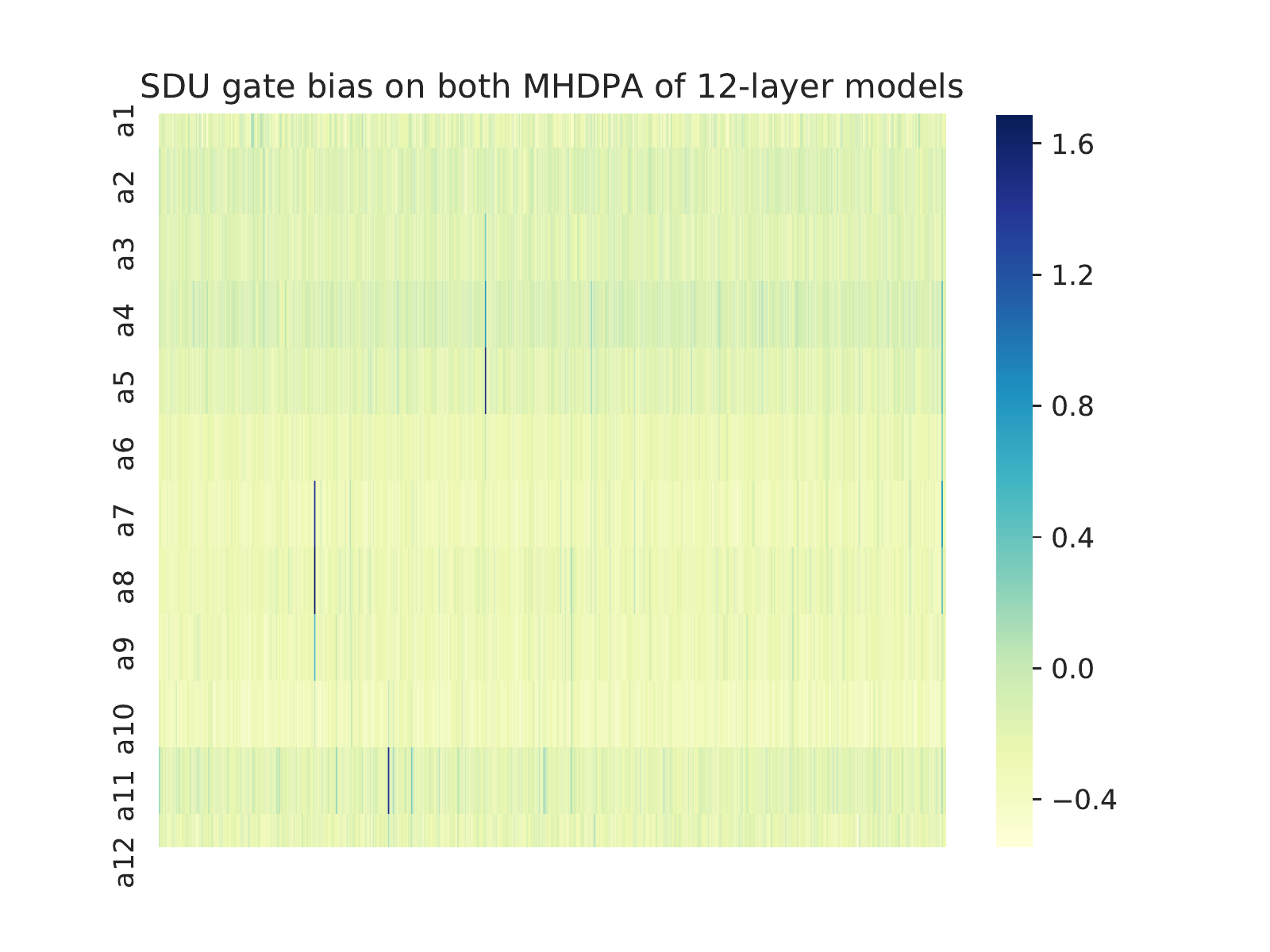}
\caption{Plot of gate biases on MHDPA of 12-layer models.} \label{ap-fig:L12-heat-b-sa}
\end{subfigure}

\medskip
\begin{subfigure}{0.48\textwidth}
\includegraphics[width=\linewidth]{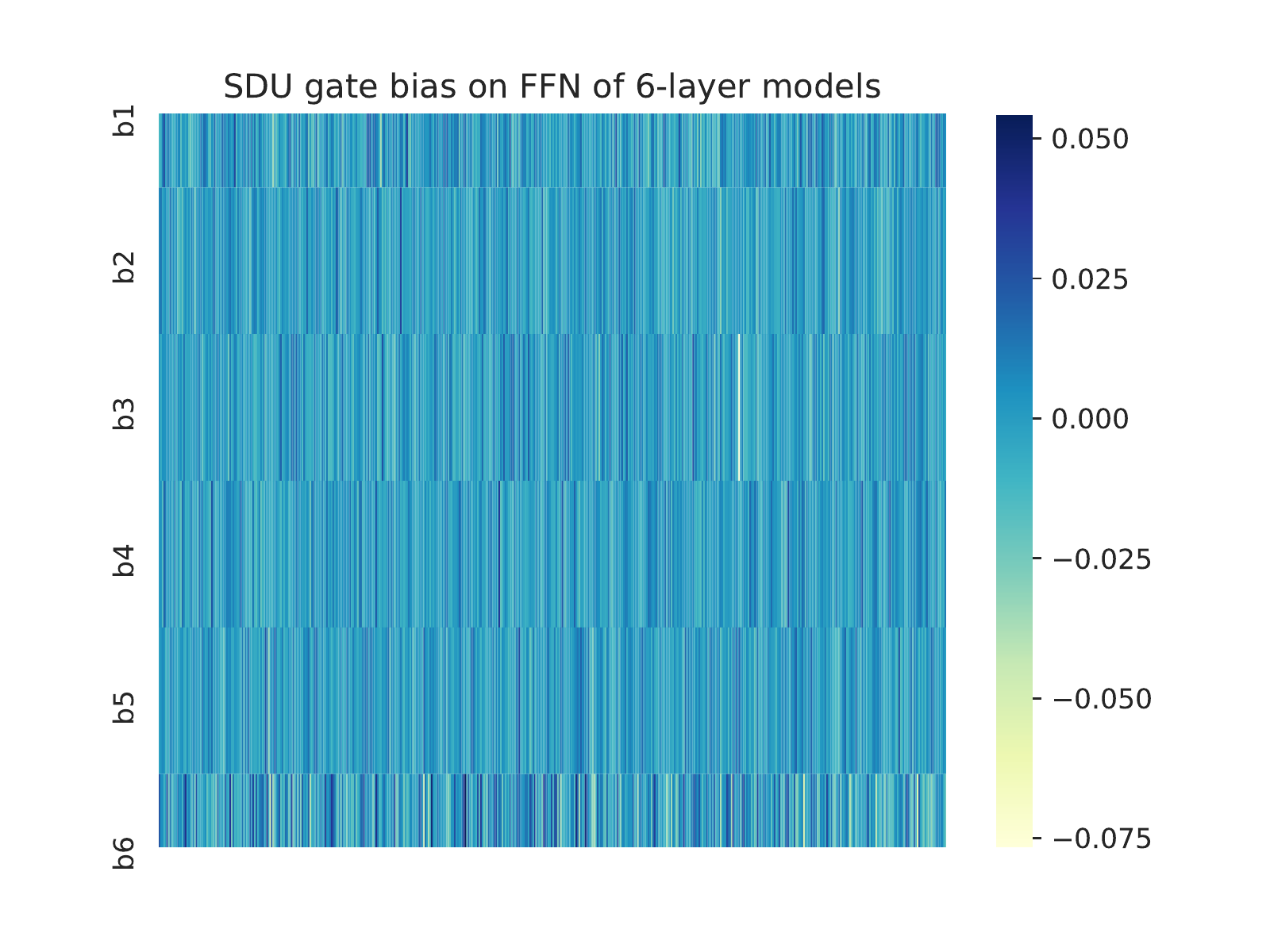}
\caption{Plot of gate biases on FFN of 6-layer models.} \label{ap-fig:L6-heat-b-ffn}
\end{subfigure}\hspace*{\fill}
\begin{subfigure}{0.48\textwidth}
\includegraphics[width=\linewidth]{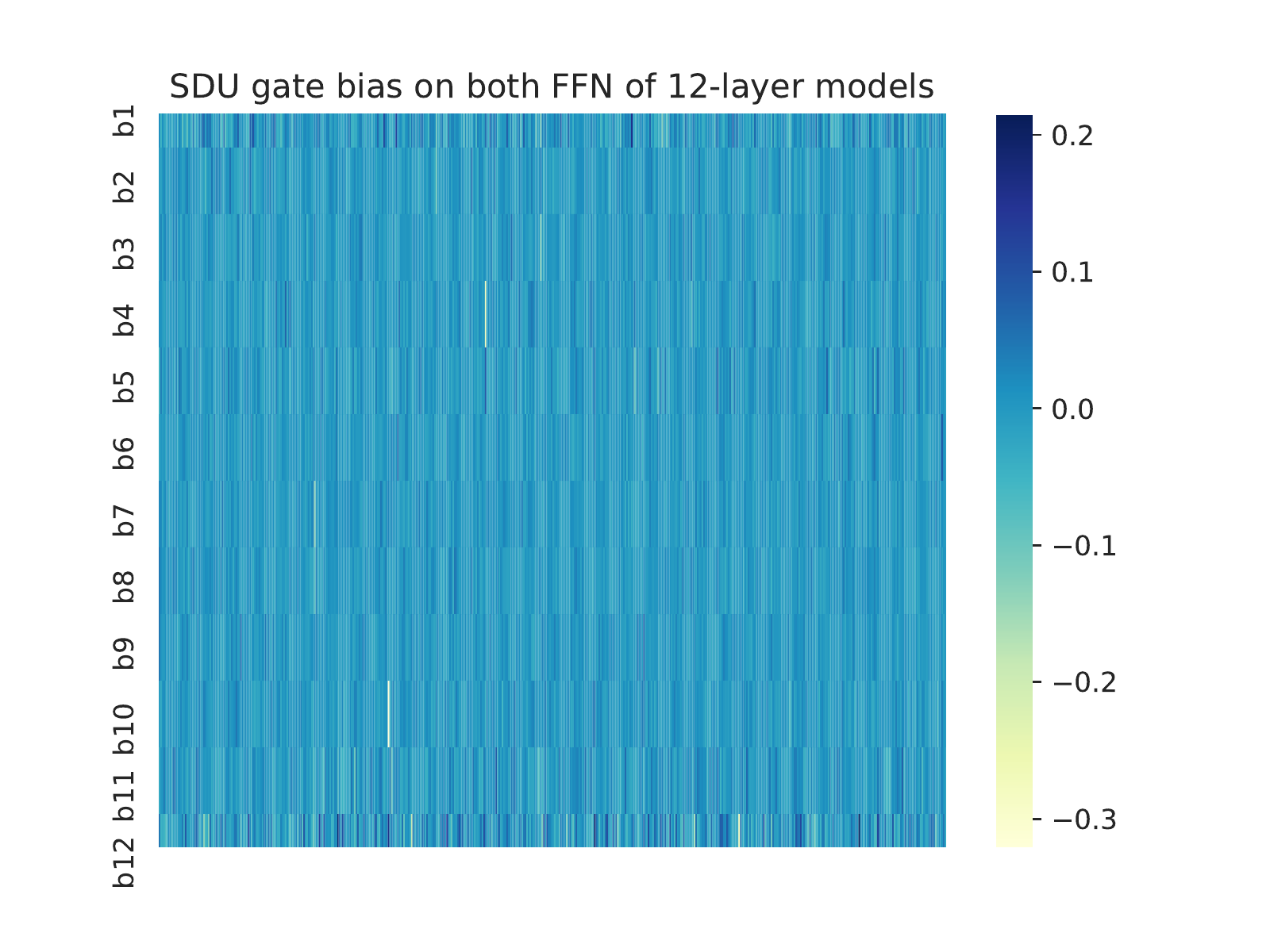}
\caption{Plot of gate biases on FFN of 12-layer models.} \label{ap-fig:L12-heat-b-ffn}
\end{subfigure}

\medskip
\begin{subfigure}{0.48\textwidth}
\includegraphics[width=\linewidth]{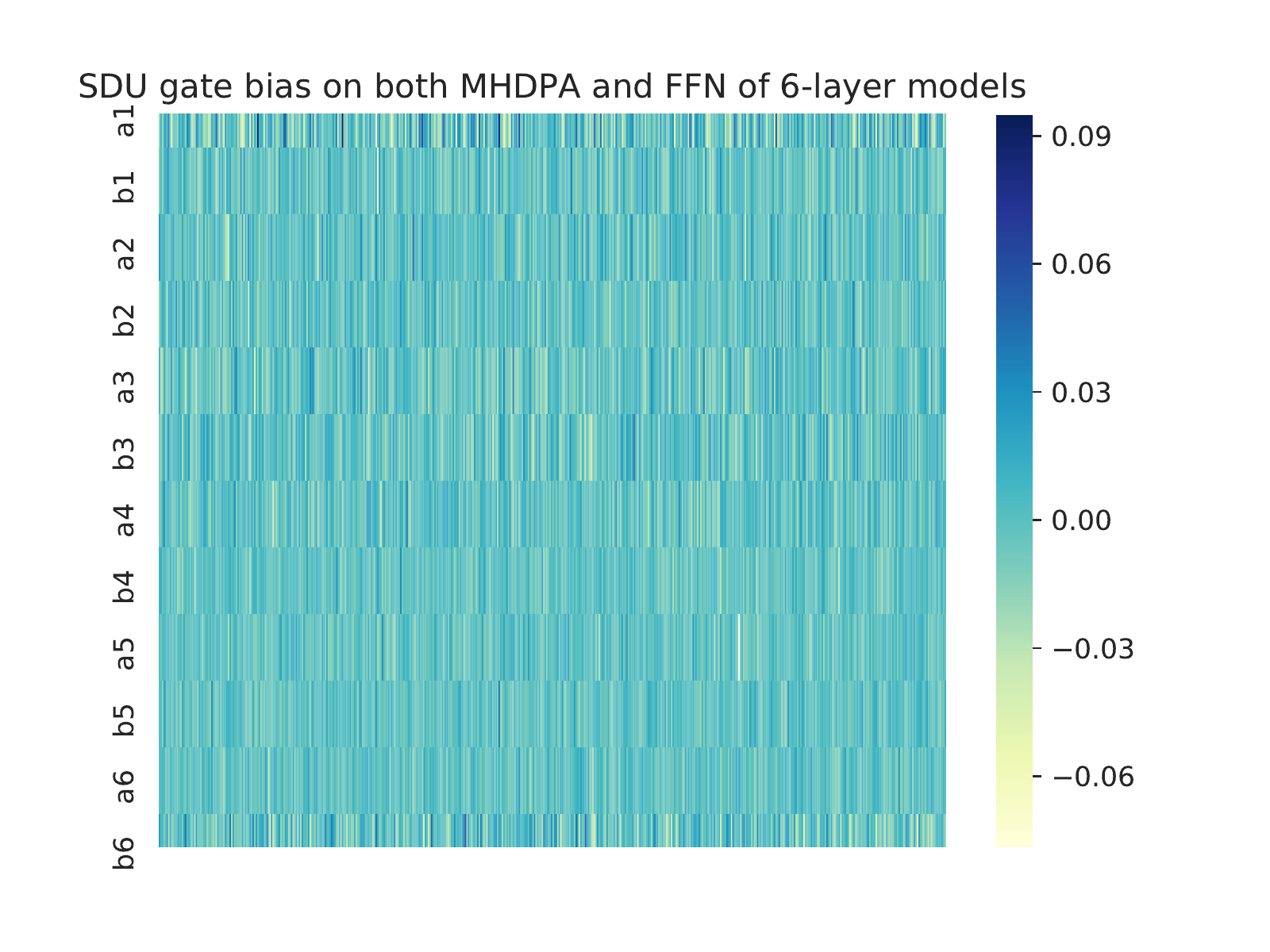}
\caption{Plot of gate biases on all sublayers of 6-layer models.} \label{ap-fig:L6-heat-b-all}
\end{subfigure}\hspace*{\fill}
\begin{subfigure}{0.48\textwidth}
\includegraphics[width=\linewidth]{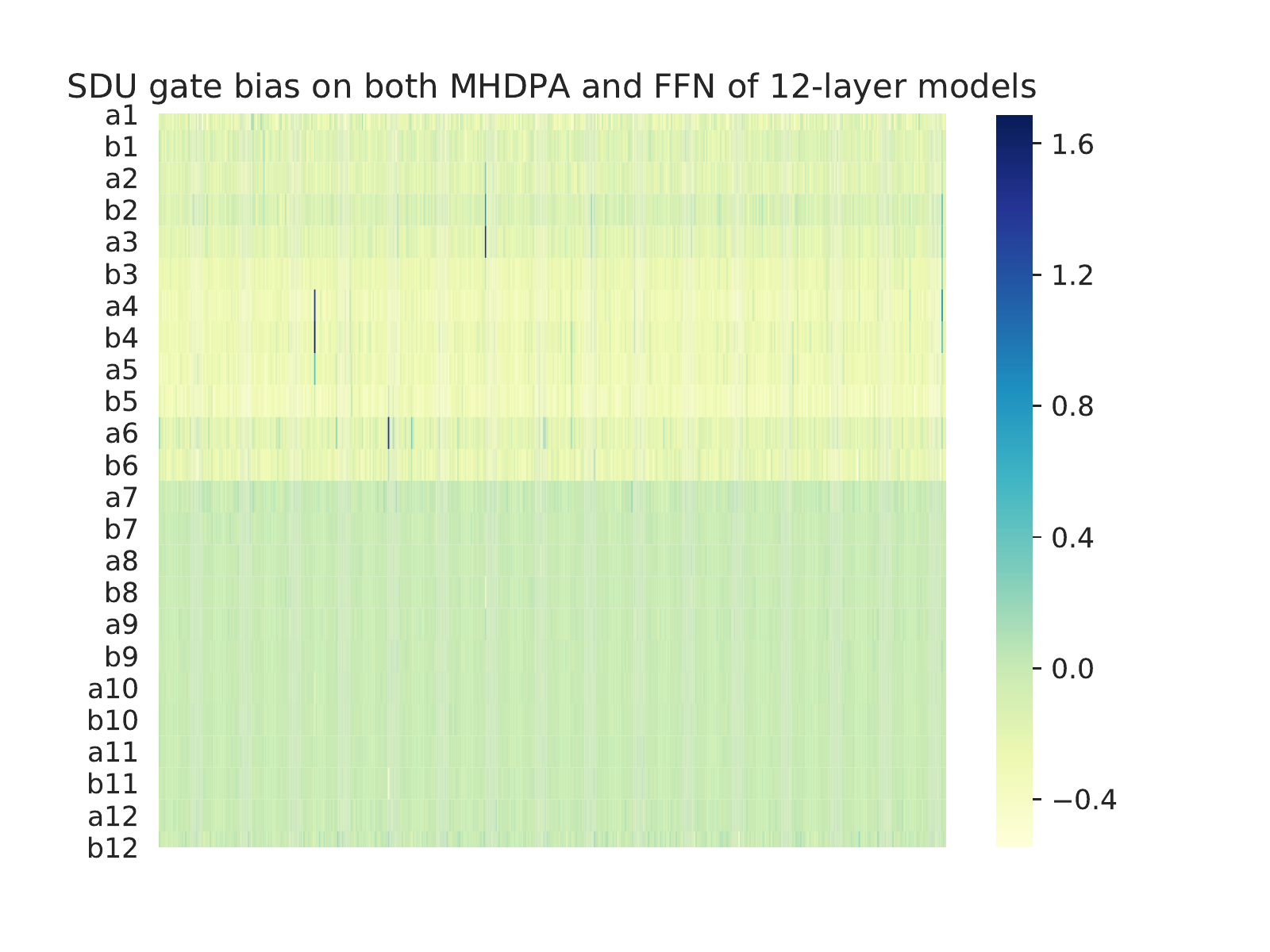}
\caption{Plot of gate biases on all sublayers of 12-layer models.} \label{ap-fig:L12-heat-b-all}
\end{subfigure}

\caption{The heatmap visualization of \textbf{learnable biases} (i.e., $\mathbf{b_1}$ in Eq.~\ref{eq:sdu-gate}) on $\sigma$ gate units of 6-layer (left column) and 12-layer (right column) \textbf{Transformer-XL} models, where vertical axises represent the layer number of our models, and ``a1'' and ``b3'' denote the 1-st MHDPA sublayer and 3-rd FFN sublayer, respectively. All gate biases are initialized as 0s with 512 dimension of each.} \label{ap-fig:heat}
\end{figure*}

Fig~\ref{ap-fig:heat} witnesses the visualization of learned biases, which are all initialized as zeros at the beginning. Obviously, the trainable biases of SDU gates perform quite different between on MHDPA and FFN sublayers as in Fig.~\ref{ap-fig:L6-heat-b-sa},~\ref{ap-fig:L6-heat-b-ffn} for 6-layer models and Fig.~\ref{ap-fig:L12-heat-b-sa},~\ref{ap-fig:L12-heat-b-ffn} for 12-layer models. Also, the gate biases are similarly distributed on all of the 6 layers, as in Fig.~\ref{ap-fig:L6-heat-b-all}, while showing the layer separation on the bottom few transformer layers as shown in Fig.~\ref{ap-fig:L12-heat-b-all}. This also verifies the experimental evidence that SDU gates on 6-layer models all positively influence the final test performance, but those only on the previous few layers of 12-layer transformers could have better results on both convergence speed and the final test bpc.

\subsection{Scatter Visualization}
\label{ap-sec:scatter}
\begin{figure*}[t]\vskip -5mm
     \centering
     \begin{subfigure}[b]{0.8\textwidth}
         \centering
         \includegraphics[width=\textwidth]{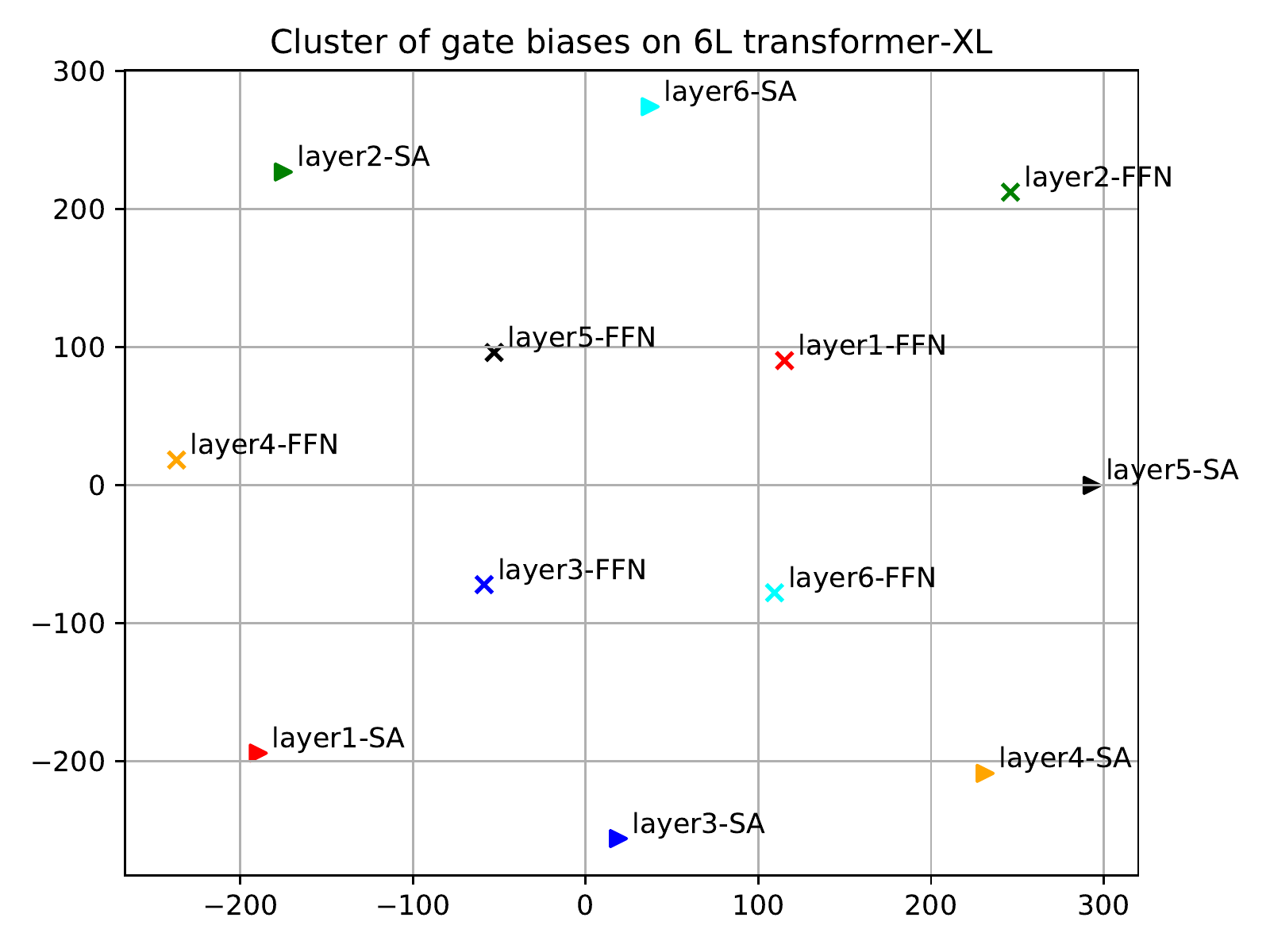}
         \caption{Plot of bias distributions on 6-layer models.}
         \label{ap-fig:L6-b-cluster}
     \end{subfigure}
     \hfill
     \begin{subfigure}[b]{0.8\textwidth}
         \centering
         \includegraphics[width=\textwidth]{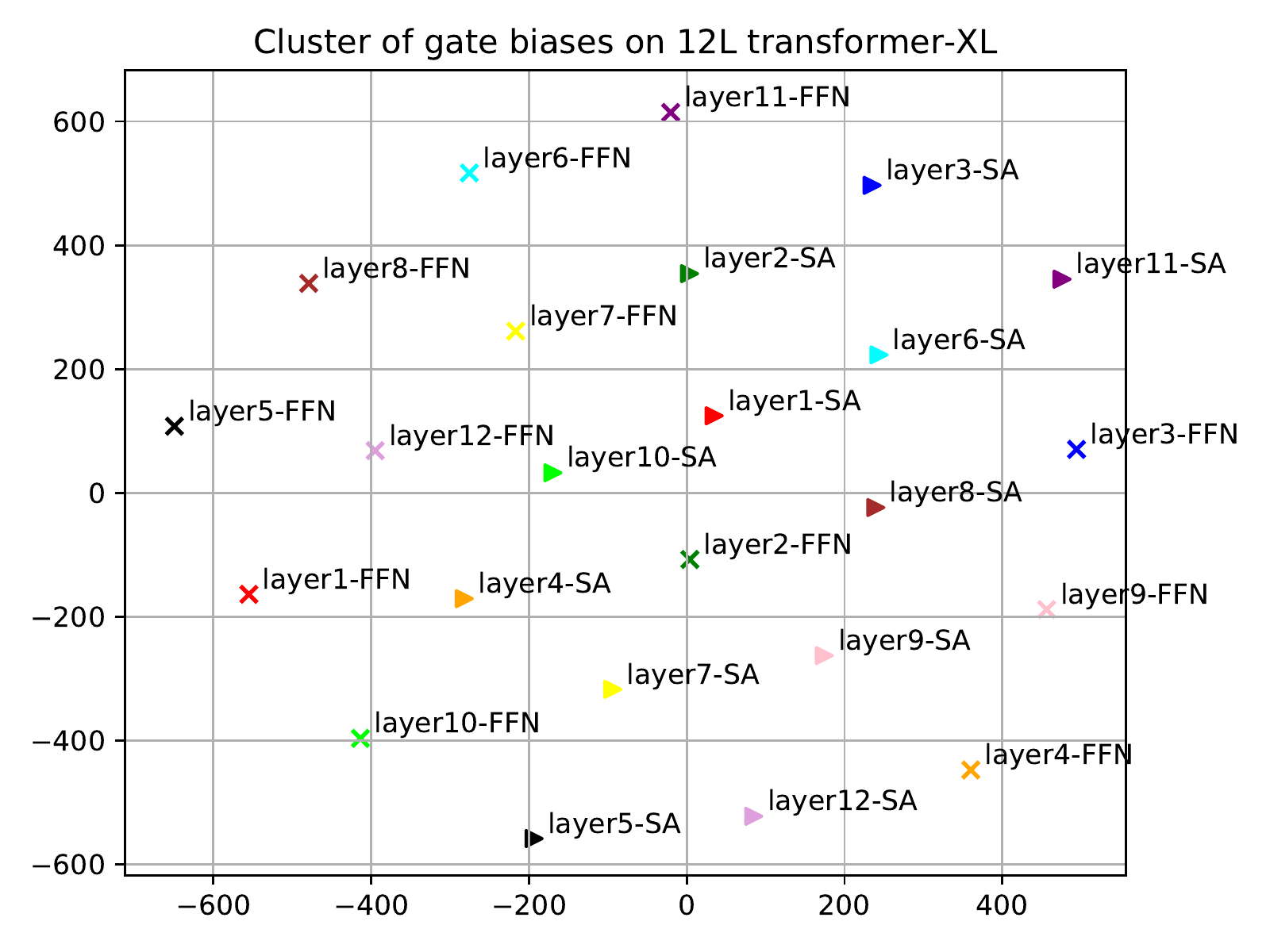}
         \caption{Plot of bias distributions on 12-layer models.}
         \label{ap-fig:L12-b-cluster}
     \end{subfigure} \vskip -2mm
        \caption{Scatter visualization of SDU gate biases on 6-layer and 12-layer Transformer-XL, where ``layer2-SA'' denotes the gate bias on 2-nd self-attention sublayer. We employ t-Distributed Stochastic Neighbor Embedding (t-SNE) to reduce the dimension from 512 to 2. }
        \label{ap-fig:cluster-xl}
        \vskip -4mm
\end{figure*}

Fig.~\ref{ap-fig:cluster-xl} illustrates the uniform distribution on both 6-layer and 12-layer Transformer-XL models. Due to the existence of residual connections, the representation space can be seen as the same. Hence the evenly distributed gate biases may learn from different aspects accordingly, which also matches our common intuition.

\clearpage

% \section{Supplemental Material}
% \label{sec:supplemental}

\end{document}